\pdfoutput=1

\documentclass[11pt]{article}

\usepackage{emnlp2022}

\usepackage{times}
\usepackage{latexsym}
\usepackage{multirow}
\usepackage{sidecap}
\usepackage{amsthm}
\usepackage{amsmath}
\usepackage{amssymb}
\usepackage{amsfonts}
\usepackage{verbatim} 
\usepackage{url}
\usepackage{pifont}
\usepackage{eqparbox}
\usepackage{arydshln} 
\usepackage{bigstrut}
\usepackage[linewidth=1pt]{mdframed}
\usepackage{framed}
\usepackage{xcolor,colortbl}
\usepackage{pgffor}
\usepackage{comment}
\usepackage{graphicx}
\usepackage{booktabs}
\usepackage{subcaption}
\usepackage{enumitem}
\usepackage{makecell}
\usepackage{boxedminipage}
\usepackage{xspace}
\usepackage{array}
\usepackage{framed}
\usepackage{caption}
\usepackage{subcaption}
\usepackage{framed}
\usepackage{array}
\newcolumntype{L}[1]{>{\raggedright\let\newline\\\arraybackslash\hspace{0pt}}m{#1}}
\newcolumntype{C}[1]{>{\centering\let\newline\\\arraybackslash\hspace{0pt}}m{#1}}
\newcolumntype{R}[1]{>{\raggedleft\let\newline\\\arraybackslash\hspace{0pt}}m{#1}}

\usepackage[T1]{fontenc}

\usepackage[utf8]{inputenc}

\usepackage{microtype}
\usepackage{soul}

\definecolor{lightergray}{RGB}{230,230,230}
\definecolor{DarkRed}{RGB}{130,25,0}
\definecolor{DarkGreen}{RGB}{30,130,30}

\newcommand{\name}{\textsc{Super-NaturalInstructions}}
\newcommand{\nameShort}{\textsc{Sup-NatInst}}
\newcommand{\modelName}{\textsc{T$k$-Instruct}}
\newcommand{\naturalins}{\textsc{NaturalInstructions}}
\newcommand{\naturalinsShort}{\textsc{NatInst}}
\newcommand{\crossfit}{\textsc{CrossFit}}
\newcommand{\flan}{\textsc{FLAN}}
\newcommand{\gptthree}{\textsc{InstructGPT}}

\newcommand{\promptsource}{\textsc{PromptSource}}
\newcommand{\bigbench}{\textsc{BigBench}}

\newcommand{\daniel}[1]{\textcolor{blue}{[DK: #1]}}

\newcommand{\yizhong}[1]{\textcolor{violet}{[YZ: #1]}}

\newcommand{\todo}[1]{{\color{red} [TODO: {#1}]}}
\newcommand{\cmark}{\textcolor{DarkGreen}{\ding{51}}}
\newcommand{\xmark}{\textcolor{red}{\ding{55}}}%

\title{\name: A Benchmark of 1,600+ Language Tasks and Instructions}
\title{A Benchmark of 1,600+ Language Tasks and Instructions}
\title{\name: A Benchmark of \\ 1,600+ Language Tasks and Instructions}
\title{\name: 1,600+ Language Tasks and Instructions}
\title{Scaling Cross-Task Generalization via ural Language Instructions}
\title{1,600+ Language Tasks for Benchmarking Instructional In-Context Learning}
\title{Benchmarking Generalization via In-Context Instructions \\  on 1,600+ Language Tasks}
\title{\name: Benchmarking Generalization via Instructions on 1,600+ NLP Tasks}
\title{Generalization via Instructions on 1,600+ NLP Tasks in \textsc{{\normalsize Super}NaturalInstructions}}
\title{\textsc{$^\mbox{\normalsize Super}$NaturalInstructions}:  \\
Generalization via Declarative Instructions at Scale}
\title{
\textsc{\name}:  \\
Generalization via Declarative Instructions on 1600+ NLP Tasks}

\author{
\fontsize{10.2pt}{10.2pt}\selectfont
 \makecell{$^\diamondsuit$Yizhong Wang$^{2}$  \; $^\diamondsuit$Swaroop Mishra$^{3}$  \; $^\clubsuit$Pegah Alipoormolabashi$^{4}$  \; $^\clubsuit$Yeganeh Kordi$^{5}$ \\
 Amirreza Mirzaei$^4$ \; Anjana Arunkumar$^3$  \; Arjun Ashok$^6$  \; Arut Selvan Dhanasekaran$^3$   \\ 
 Atharva Naik$^{7}$ \; David Stap$^{8}$  \; Eshaan Pathak$^{9}$  \; Giannis Karamanolakis$^{10}$  \;  Haizhi Gary Lai$^{11}$ \\ 
 Ishan Purohit$^{12}$ \; Ishani Mondal$^{13}$  \; Jacob Anderson$^{3}$  \; Kirby Kuznia$^{3}$ \; Krima Doshi$^{3}$  \; Maitreya Patel$^{3}$  \\  
  Kuntal Kumar Pal$^{3}$ \; Mehrad Moradshahi$^{14}$ \; Mihir Parmar$^{3}$ \; Mirali Purohit$^{15}$ \; Neeraj Varshney$^{3}$ \\ 
  Phani Rohitha Kaza$^{3}$ \; Pulkit Verma$^{3}$ \; Ravsehaj Singh Puri$^{3}$ \; Rushang Karia$^{3}$ \; Shailaja Keyur Sampat$^{3}$  \\ 
  Savan Doshi$^{3}$ \; Siddhartha Mishra$^{16}$ \; Sujan Reddy$^{17}$  \; Sumanta Patro$^{18}$ \; Tanay Dixit$^{19}$ \; Xudong Shen$^{20}$ \\ 
Chitta Baral$^{3}$ \; Yejin Choi$^{1,2}$  \; Noah A. Smith$^{1,2}$\; Hannaneh Hajishirzi$^{1,2}$  \; Daniel Khashabi$^{21}$
}
\vspace{2pt}
\\
\fontsize{7.1pt}{7.1pt}\selectfont
\makecell{
$^{1}$Allen Institute for AI
$^{2}$Univ. of Washington
$^{3}$Arizona State Univ.
$^{4}$Sharif Univ. of Tech.
$^{5}$Tehran Polytechnic
$^{6}$PSG College of Tech.
$^{7}$IIT Kharagpur \\ 
$^{8}$Univ. of Amsterdam
$^{9}$UC Berkeley
$^{10}$Columbia Univ.
$^{11}$Factored AI
$^{12}$Govt. Polytechnic Rajkot
$^{13}$Microsoft Research
$^{14}$Stanford Univ.   
$^{15}$Zycus Infotech \\
$^{16}$Univ. of Massachusetts Amherst
$^{17}$National Inst. of Tech. Karnataka
$^{18}$TCS Research
$^{19}$IIT Madras
$^{20}$National Univ. of Singapore
$^{21}$Johns Hopkins Univ.
}
}

\newcommand\blfootnote[1]{%
  \begingroup
  \renewcommand\thefootnote{}\footnote{#1}%
  \addtocounter{footnote}{-1}%
  \endgroup
}

\begin{document}
\maketitle

\begin{abstract}

How well can NLP models generalize to a \emph{variety} of unseen tasks when provided with task instructions?
To address this question, we first introduce \name,\footnote{\name{} represents a super-sized expansion of \naturalins{}~\cite{mishra2022cross} which had 61 tasks.} a benchmark of 1,616 diverse NLP tasks and their expert-written instructions. 
 Our collection covers 76 distinct task types, 
 including but not limited to classification, extraction, infilling, sequence tagging, text rewriting, and text composition.
This large and diverse collection of tasks enables rigorous benchmarking of cross-task generalization under instructions---training models to follow instructions on a subset of tasks and evaluating them on the remaining unseen ones.
\\
Furthermore, we build \modelName{}, a transformer model 
trained to follow a variety of in-context instructions (plain language task definitions or $k$-shot examples). 
Our experiments show that \modelName{}  outperforms existing 
instruction-following models such as InstructGPT by over 9\% on our benchmark despite being an order of magnitude smaller. 
We further analyze generalization as a function of various scaling parameters, such as the number of observed tasks, the number of instances per task, and model sizes.
We hope our dataset and model facilitate future progress towards more general-purpose 
NLP models.\footnote{
The dataset, models, and 
a leaderboard
can be found at \url{https://instructions.apps.allenai.org}. 
\label{footnote:1}
}
\blfootnote{$\diamondsuit$ Co-first authors $\clubsuit$ Co-second authors}

\end{abstract}


\begin{figure}[t]
    \centering
    \includegraphics[width=\columnwidth,trim=0cm 0cm 0cm 1cm]{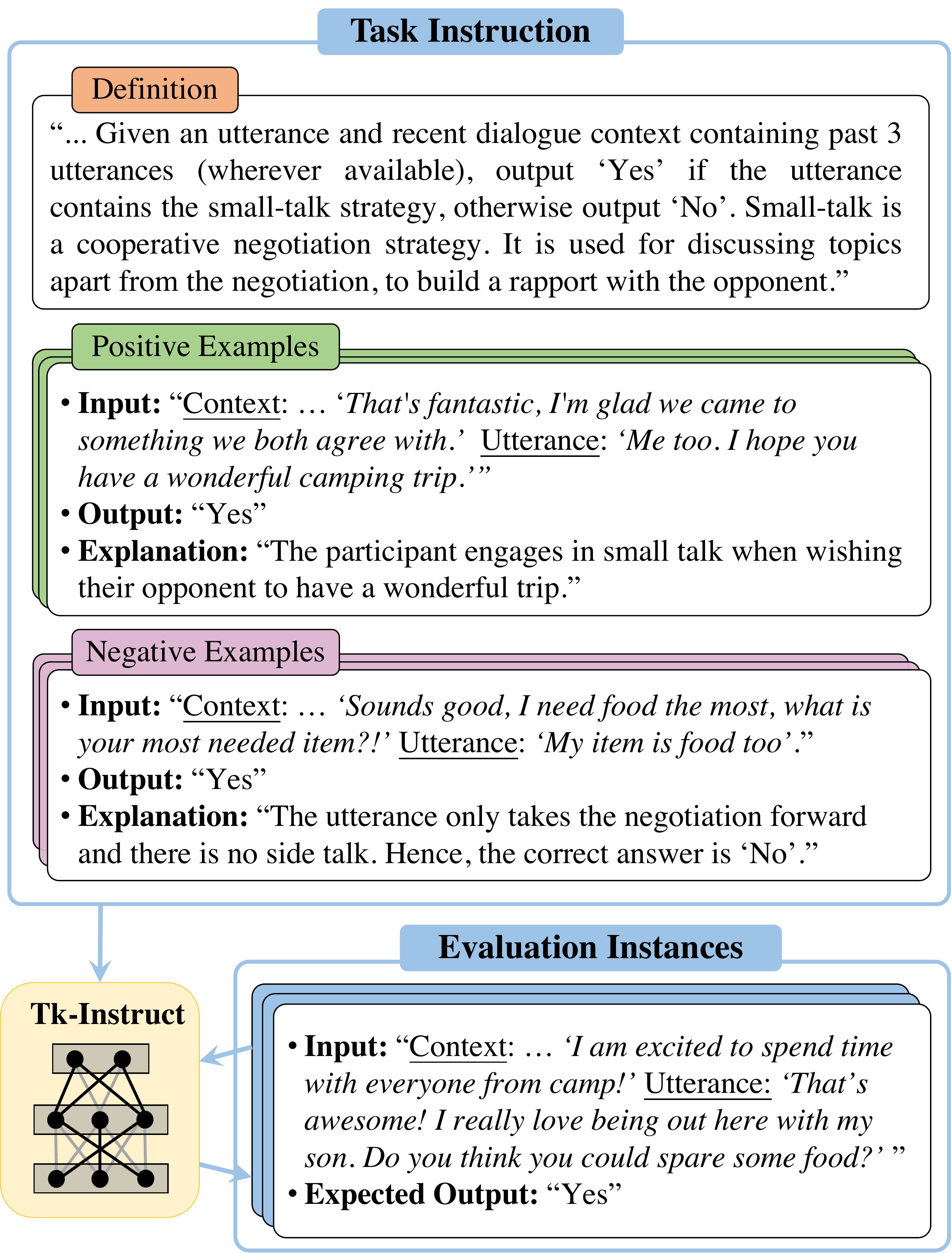}
    \caption{An example task from \nameShort{} adopted from \citet{chawla2021casino}. 
    A successful model is expected to use the provided instructions (including task definition and demonstration examples) to output responses to a pool of evaluation instances. 
  }
    \label{fig:example}
\end{figure}

\newcommand{\unk}{{\small \texttt{unk}}}

\begin{table*}[ht]
    \centering
    \small 
    \resizebox{\linewidth}{!}{
    \begin{tabular}{lcccccc}
        \toprule
        Resource → & \makecell{ \nameShort \\(this work)} & \makecell{\naturalinsShort \\\cite{mishra2022cross}} & 
        \makecell{\crossfit \\ \cite{ye2021crossfit}} &  \makecell{\promptsource\\\cite{bach2022promptsource}} & \makecell{\flan\\\cite{wei2022finetuned}} & \makecell{\gptthree \\\cite{ouyang2022training}} 
        \\
         \cmidrule(lr){1-1}  \cmidrule(lr){2-2}  \cmidrule(lr){3-3}  \cmidrule(lr){4-4}  \cmidrule(lr){5-5} \cmidrule(lr){6-6}  \cmidrule(lr){7-7} 
          Has task instructions?   & \cmark & \cmark & \xmark & \cmark & \cmark & \cmark  \\ 
          Has negative examples?   & \cmark & \cmark & \xmark & \xmark & \xmark & \xmark  \\ 
          Has non-English tasks? & \cmark & \xmark &  \xmark & \xmark & \cmark & \cmark   \\ 
          Is public?   & \cmark & \cmark & \cmark & \cmark & \cmark & \xmark   \\ 
          \hdashline
        
          Number of tasks & 1616 & 61 & 269 & 176 & 62 & -- \\
          Number of instructions & 1616 & 61 & -- & 2052 & 620 & 14378 \\
          Number of annotated tasks types & 76 & 6 & 13 & 13$^*$ & 12 & 10 \\
          Avg.~task definition length (words)  & 56.6 & 134.4 & -- & 24.8 & 8.2 & --  \\ 
         \bottomrule 
    \end{tabular}
    }
    \caption{
        A comparison of \nameShort{} to a few notable datasets in the field. We obtain the number of tasks, instructions, and task types of other datasets from their original paper. ``--'' indicates the fields are not applicable or unknown.  
        Standards for categorizing task types vary across different datasets (see Fig. \ref{fig:visualizations}). *\promptsource~does not provide task type annotation for all their tasks, for which we report only the 13 task types annotated for training T0 \cite{sanh2022multitask} instead. 
    }
    \label{tab:dataset_comparison}
\end{table*}

\begin{figure*}
\centering
\begin{tabular}{cccc}
\multicolumn{2}{c}{
\multirow{2}{*}{
    \begin{minipage}[b][][t]{0.4\textwidth}\vspace{0pt}
	\begin{subfigure}[b][1\width]{1\textwidth}
	   \includegraphics[trim= 2cm 0.3cm 0cm 3cm, width=1.14\textwidth]{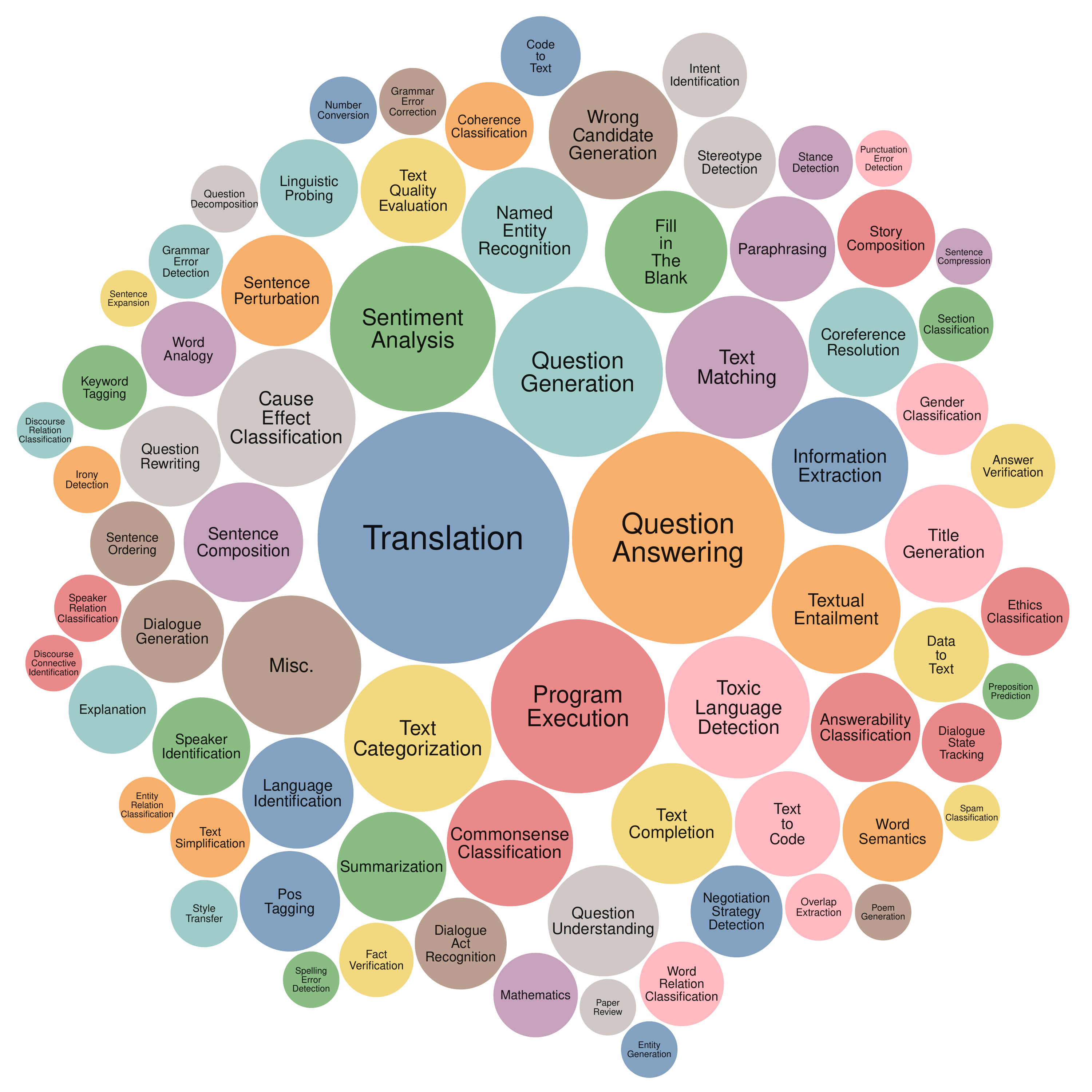}
	   \captionsetup{font=scriptsize}\caption{\nameShort{} (this work) \label{figure:visualization-nli2}}
	\end{subfigure}
	\end{minipage}
}} & \begin{subfigure}[c][1\width]{
	   0.22\textwidth}
	   \includegraphics[width=1\textwidth]{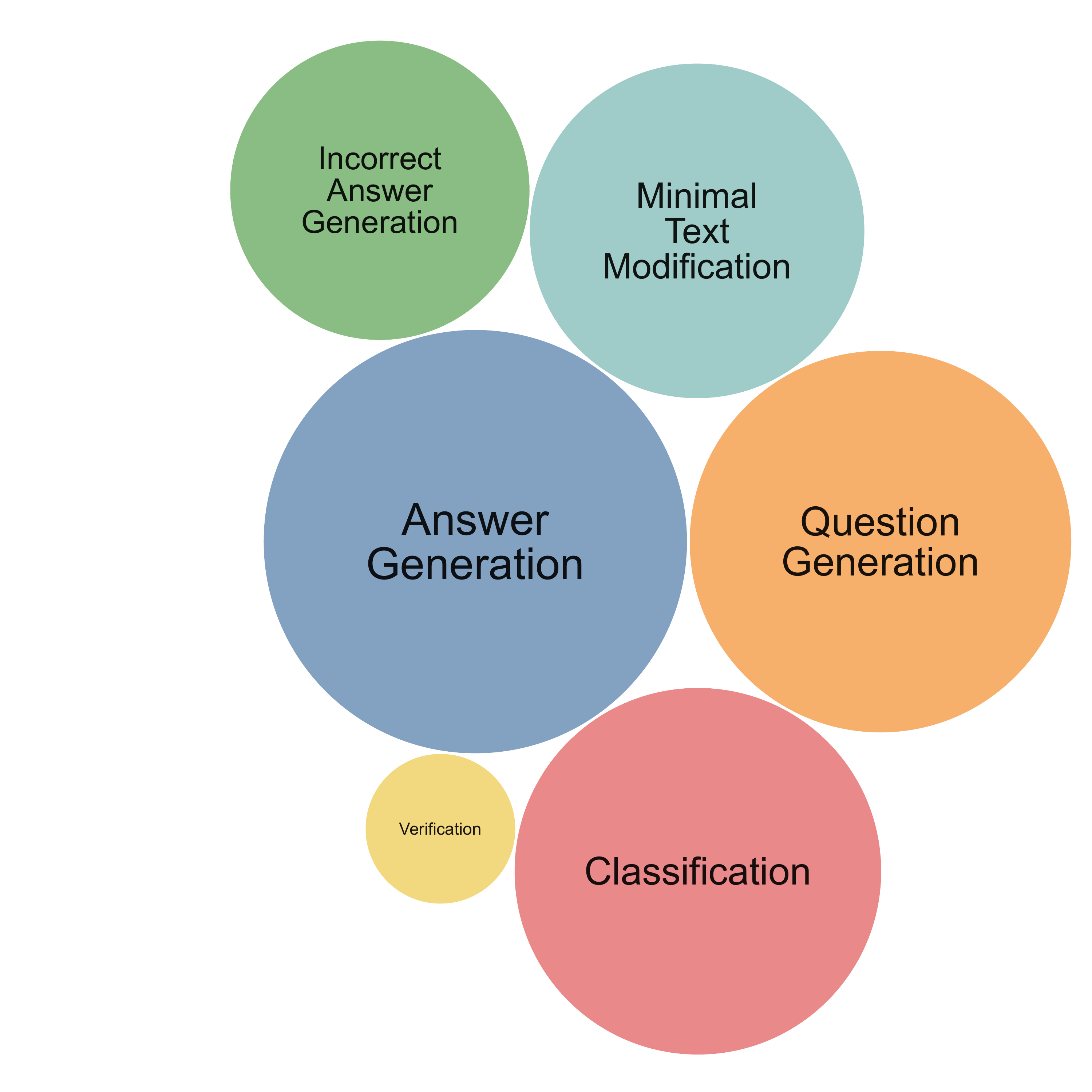}
	   \captionsetup{font=scriptsize}\caption{\naturalinsShort \label{figure:visualization-nli1}}
	\end{subfigure} &
	\begin{subfigure}[c][1\width]{
	   0.22\textwidth}
	   \includegraphics[width=1\textwidth]{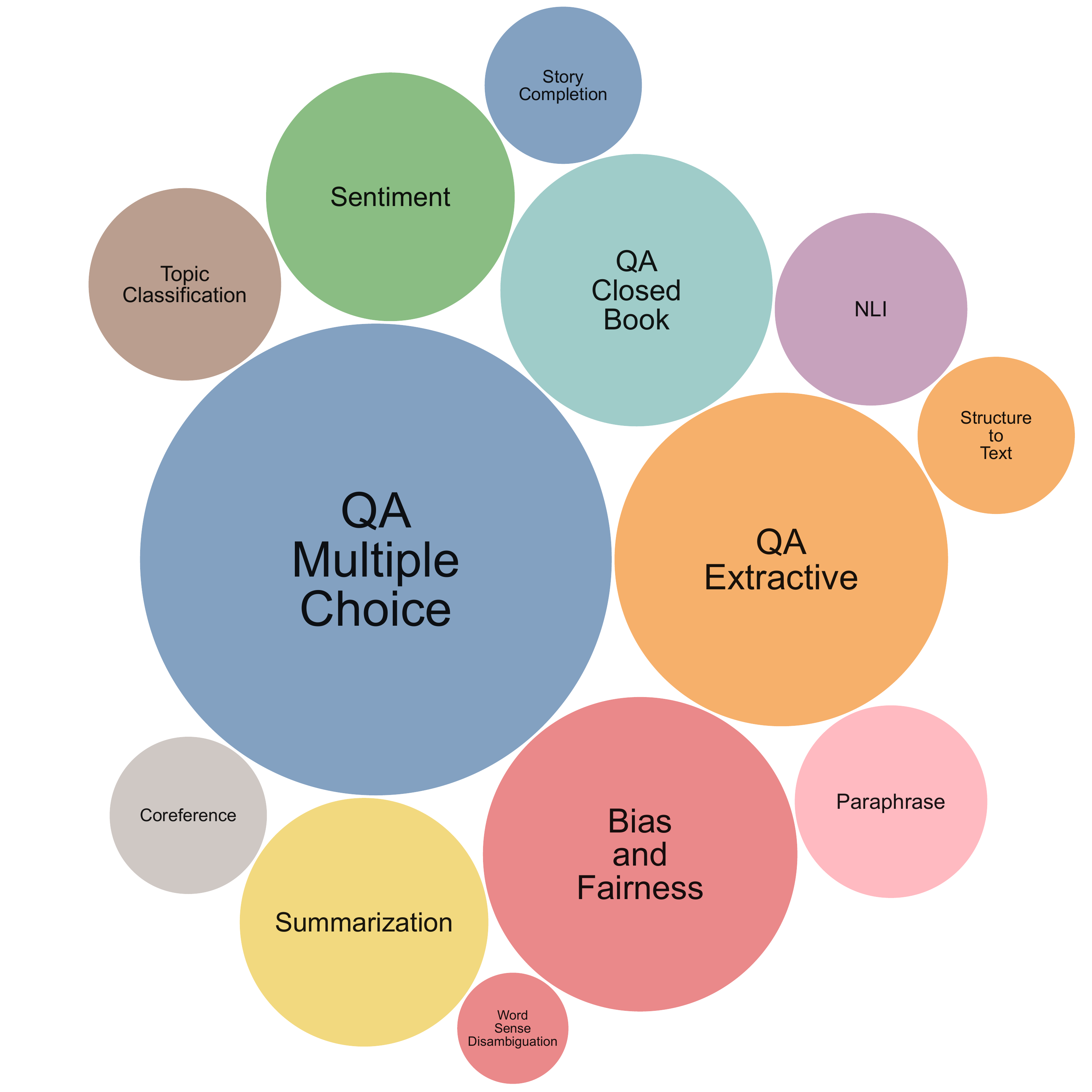}
	   \captionsetup{font=tiny}\caption{\promptsource~(T0 subset) \label{figure:visualization-ps}}
	\end{subfigure}	
	\\[2.1cm]
\multicolumn{2}{c}{} &
    \begin{subfigure}[c][1\width]{
	   0.22\textwidth}
	   \centering
	   \includegraphics[width=1\textwidth]{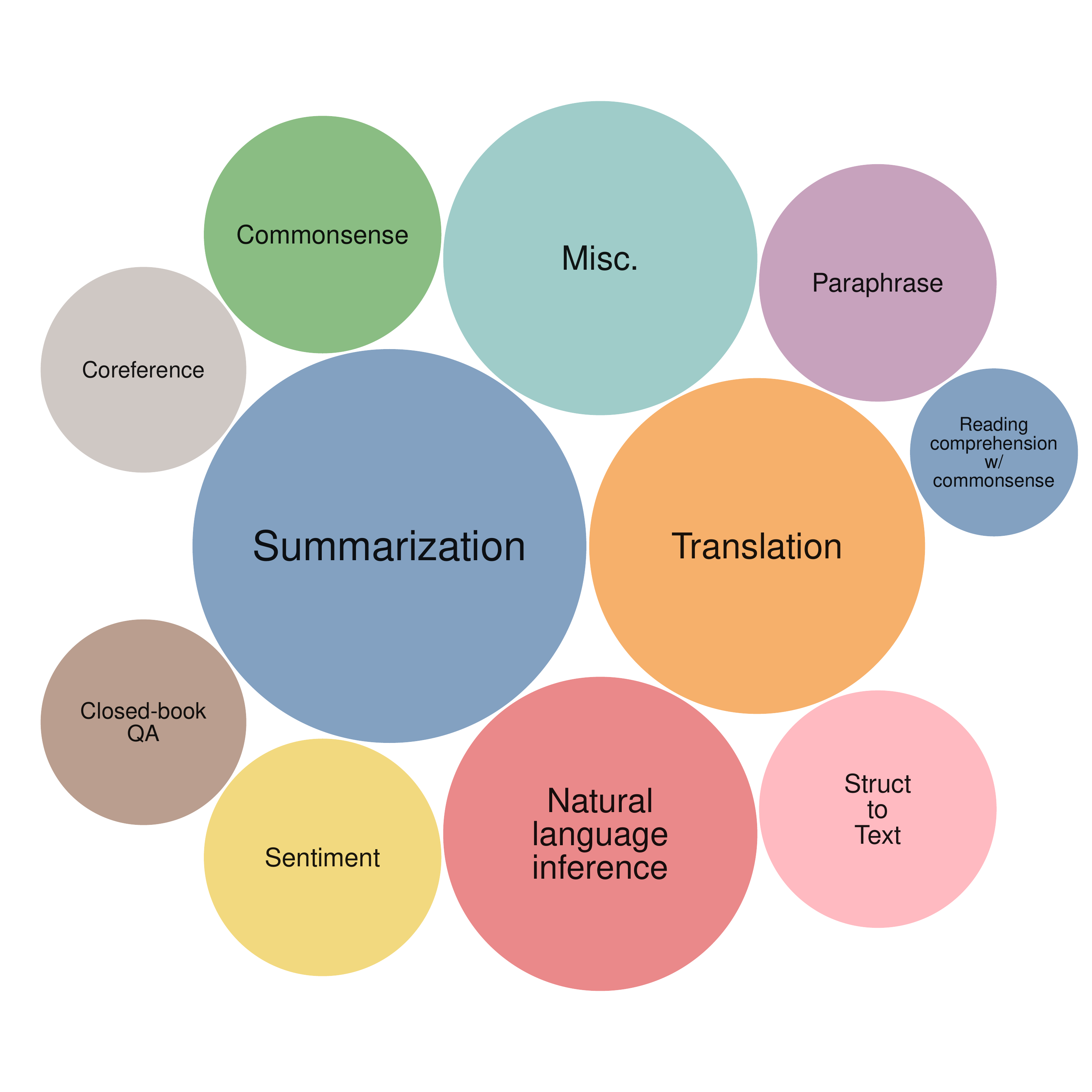}
	   \captionsetup{font=scriptsize}\caption{\flan \label{figure:visualization-flan}}
	\end{subfigure} &
	\begin{subfigure}[c][1\width]{
	   0.22\textwidth}
	   \centering
	   \includegraphics[width=1\textwidth]{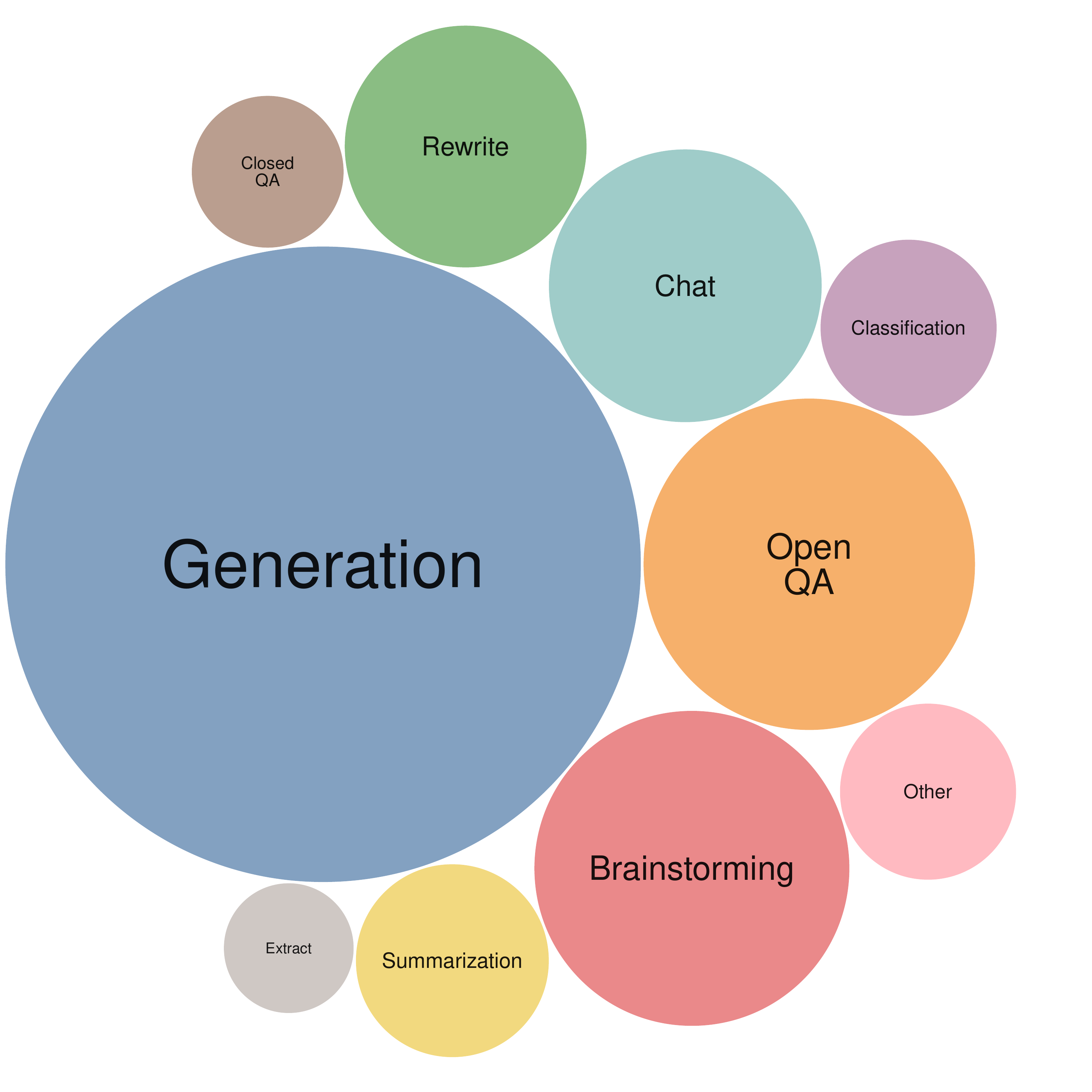}
	   \captionsetup{font=scriptsize}\caption{\gptthree \label{figure:visualization-gpt}}
	\end{subfigure} \\
	\end{tabular}
	\vspace{5pt}
    \caption{
    Compared to other datasets, \nameShort{} covers a more diverse range of task types. InstructGPT reports a very coarse categorization of their task types. Bubble size represents the number of tasks of each type in log scale.}    
    \label{fig:visualizations}
\end{figure*}

\section{Introduction}

The NLP community has witnessed great progress in building models for generalization to  {\it unseen} tasks via in-context instructions~\cite{mishra2022cross,sanh2022multitask,wei2022finetuned}
using large pretrained language models 
~\cite{raffel2020exploring,brown2020language}.
As remarkable as models like InstructGPT~\cite{ouyang2022training} are, the contribution of various design choices to their success  is  opaque. In particular, the role of supervised data has remained understudied due to limited data released by the corporate entities behind major models. In addition,  it is nearly impossible for the research community to extend  and re-train these  gigantic models. Addressing these two challenges necessitates the availability of large-scale public benchmarks of a broad range of NLP tasks and their instructions to facilitate developing and evaluating models that can generalize to unseen tasks.

In this paper, we construct a meta-dataset (i.e., dataset of datasets; \citealp{triantafillou2019meta}) that consists of a wide variety of NLP tasks with their instructions, and 
train a model that can perform a new task given the instruction,
outperforming InstructGPT (which uses $16\times$ more parameters).


Our dataset, \name{} (\nameShort{} for short), is a large benchmark of 1,616 NLP tasks and their natural language instructions. It brings in a
diverse variety of tasks---76 broad task types spanning 55 different languages.
Each task is paired up with an instruction that consists of the task definition for mapping an input text to a task output and several examples for demonstrating the desired or undesired output (see Fig.\ref{fig:example} as an example task).
These tasks and their instructions are contributed by 88 NLP practitioners, in response to our public call.
These contributions are consolidated after several rounds of peer-review and crowdsourced feedback to 
ensure quality.
Having this diverse and large-scale data enables us to carefully split the tasks into training and test sets and systematically study how state-of-the-art methods perform on them.  Table~\ref{tab:dataset_comparison} and Figure~\ref{fig:visualizations} highlight properties of \nameShort{} compared to relevant benchmarks, emphasizing the diversity of tasks and instruction types in our benchmark.


Our model, \modelName{}, is a generative model for transforming 
task inputs given declarative in-context instructions (task definition or $k$-shot examples). 
It is built by multi-task training of the T5 model~\cite{raffel2020exploring} over all the task instructions in our training set, and is evaluated on unseen tasks in the test set.  Interestingly, an 11B-parameter \modelName{} can 
outperform the 175B-parameter InstructGPT model by 9.9 ROUGE-L points when evaluated on 119 unseen English tasks, and the multilingual variant m\modelName{} outperforms InstructGPT by 13.3 points on 35 non-English tasks (\S\ref{subsec:main_results}).
According to human evaluation, \modelName{} generates responses at least as well as the ground truth for 77\% of the testing instances (\S\ref{subsec:human_eval}), confirming its strong generalization to unseen tasks. 




The compelling empirical performance of \modelName{} confirms the importance of super-sized meta datasets such as our \nameShort{} to facilitate research towards generalizable NLP models. We conduct extensive analysis to understand the important factors for this generalization (\S\ref{sec:analysis}). Our analysis shows that scaling up the diversity of training tasks and the model size are both important for strong generalization to unseen tasks. 
Finally, we estimate performance upper bounds, suggesting further room for improvement.

\section{Related Work}
\label{sec:related:work}
Language instructions are a versatile way of defining goals, which is why they have been studied in the context of a variety of applications, 
such as
instructions in grounded environments~\cite{shridhar2020alfred, stepputtis2020language,min2022film,weir2022one} and  
database commands~\cite{kim2020natural}.  
Here, we focus on applications of instructions for general NLP tasks.

Recent literature has been motivated by building models that are generalizable across a variety of NLP tasks, when prompted with either a few examples~\cite{ye2021learning,bragg2021flex} or language definitions~\cite{efrat2020turking,weller2020learning,zhong2021adapting,mishra2022cross,mishra2022reframing,parmar-etal-2022-boxbart}. 
Our work is related to the existing benchmarks in this line of work, as delineated in Table~\ref{tab:dataset_comparison} along various dimensions. 
Our benchmark  extends \naturalinsShort~\cite{mishra2022cross} with  26$\times$  more tasks and greater variety of task types (Fig.~\ref{fig:visualizations}). 
While \crossfit~\cite{ye2021crossfit} focuses on benchmarking with a few in-context examples, our benchmark also offers task instructions. 

Concurrent to our work, \promptsource{} \cite{bach2022promptsource} is another benchmark of tasks and their language instructions (prompts).
An important distinction between this benchmark and ours is the phrasing of the task definitions: while \promptsource{} task definitions are relatively concise, our task definitions 
are collected with the intention of providing a complete definition of each task and therefore are longer (24 tokens vs.~56 tokens on average; Table~\ref{tab:dataset_comparison}). 
More recently, \bigbench~\cite{srivastava2022beyond} introduces a collection of 204 tasks and also provides short task descriptions and input prefixes that can be used for prompting LMs. With little overlap to our collection of tasks, they focus more on finding challenging tasks that can be used to test different behaviors of current LMs.
Nevertheless, 
we believe that all these efforts in collecting different tasks as well as the task instructions are complementary, and the community will benefit from considering different benchmarks.
Finally, the well-adopted InstructGPT model~\cite{ouyang2022training} is partially enabled by a large dataset of prompts that are collected via various synthetic data augmentation which, unfortunately, is not publicly available. 




Beyond cross-task generalization, our benchmark can also be used to study multi-task learning more broadly, which is a longstanding goal for AI~\cite{caruana1997multitask}.
Traditionally, this literature focuses on setups that involve evaluation on tasks that are observed during training~\cite{collobert2008unified, hashimoto2017a}. 
More recent studies show promise that large-scale multi-task learning can enable strong generalization to similar tasks via unified encoding \cite{khashabi2020unifiedqa,xie2022unifiedskg} or better finetuning results on downstream tasks~\cite{mccann2018natural,aribandi2022ext5}.
Our proposed benchmark provides diverse tasks for studying multi-tasking at a massive scale.

\section{\name{}}
\label{sec:dataset}

    

\name{} is a meta-dataset \cite{triantafillou2019meta} consisting of a variety of NLP tasks (see Fig. \ref{figure:visualization-nli2}) and instructions that describe them in plain language. 


\vskip .1cm \noindent {\bf Instruction schema.}
\label{schema}
All  task instructions follow the same uniform schema (see Fig.~\ref{fig:example}) which 
is composed of the following parts:

    

\begin{itemize}[noitemsep,topsep=0pt,parsep=3pt,leftmargin=0.3cm]
    \item \underline{\textsc{Definition}} defines a given task in natural language. This is 
    a complete definition of how an input text (e.g., a sentence or a document) is expected to be mapped to an output text. 
    \item \underline{\textsc{Positive Examples}} are samples of inputs and their \emph{correct} outputs, along with a short explanation for each.
    \item \underline{\textsc{Negative Examples}} are samples of inputs and their \emph{incorrect/invalid} outputs, along with a short explanation for each.
\end{itemize}
The above schema is based on that of \citet{mishra2022cross}, though it is simplified. See Appendix~\ref{sec:comparison:schema} for the comparison.

\vskip .1cm \noindent {\bf Task instances.}
Given the instructions for each task, a model is expected to solve instances of that task.
We use a unified format to organize the instances of all our tasks. 
More precisely, each instance consists of a textual input and a list of acceptable textual outputs. 
We limit the number of instances in each task to $6.5$K to avoid an imbalance of instances between tasks. 

\vskip .1cm \noindent {\bf Benchmark collection.}
The benchmark was collected through a large community effort on 
GitHub.\footnote{\url{https://github.com/allenai/natural-instructions}}
Tasks were collected and contributed by NLP practitioners who were either responding to our public invitation\footnote{\url{https://blog.allenai.org/9d3f24d5a9db}}
or students who were encouraged to contribute as part of their class project.\footnote{CSE 576 ``Topics in NLP'' course, Arizona State Univ.}
Contributors were encouraged to be creative and 
source the tasks from several resources:
(a)  existing public NLP datasets, (b) 
available intermediate annotations in crowdsourcing experiments 
(e.g., paraphrasing questions or 
rating their quality during crowdsourcing a QA dataset),
or 
(c) synthetic tasks that can be communicated to an average human in a few sentences 
(e.g., basic algebraic operations like number comparison, finding the longest palindrome substring, etc.). 
When using existing datasets or crowdsourcing annotations, contributors were encouraged to adopt the instructions used to create this dataset whenever available. 
This was done to ensure that the instructions were sufficient to define the tasks to average human readers. 
Tasks along with instructions and other meta information were contributed as JSON files via GitHub pull requests, which were reviewed by automated checks and peers. 
We had 88 contributors from diverse locations and backgrounds contribute to our repository.

\vskip .1cm \noindent {\bf Quality control.}
Controlling the quality of this 
community-contributed data
was done in several phases: 
(1) Upon creating a GitHub pull request of the proposed task, it immediately went through an automatic test. 
This process verified that the introduced file contained the expected fields and adhered to our desired properties (e.g., no duplicate instances, the output labels are not heavily imbalanced, etc.) and
(2) The proposed task
was then peer-reviewed by 1--2 other expert contributors to ensure the clarity and sufficiency 
of instruction content. 
The review process was done iteratively until the reviewers were content with the quality of the proposed instruction.
Specifically, reviewers were asked to verify that the instruction is clear and sufficient for an average language speaker to solve the underlying task (evaluation instances) while being grammatical, fluent, and concise. 
On average, the review of each GitHub pull request took about 4--6 iterations over the span of multiple days before being merged.
(3) Lastly, the added tasks were presented to crowdworkers in order to collect feedback on the quality of the provided instructions, such as typos, clarity, or other issues
(details in \S\ref{sec:crowdsourced:feedback}). 
Subsequently, one of the authors used this feedback to improve the task definitions of the instances. 
This feedback was done only for English tasks, as finding high-quality crowdworkers in other languages is nontrivial~\cite{pavlick2014language}.  


\vskip .1cm \noindent {\bf Diversity of tasks.}
Collecting tasks for \nameShort{} was carefully supervised to cover a wide variety of natural language understanding tasks, domains, and languages. 
To better understand this diversity, we comprehensively categorize tasks along three different dimensions:

\begin{itemize}[noitemsep,topsep=0pt,parsep=3pt,leftmargin=0.3cm]
    \item \underline{\textsc{Task Type}} defines the nature of the mapping from instance inputs to outputs (e.g., question answering, classification, etc.).
    \item \underline{\textsc{Language}} indicates the language(s) of the instances.
    \item \underline{\textsc{Domain}} indicates the domain(s) to which the text of the tasks belong to (e.g., politics, medicine, dialogue, etc.).
\end{itemize}
These different measures of categorization can be used to study different senses of generalization. 
In our empirical studies (\S\ref{sec:empirical}), we study generalization along the axis of task types. 
We refer the reader to Fig.~\ref{fig:histograms} in the appendix for the distribution of tasks among different task types, languages, and domains. 



\vskip .1cm \noindent {\bf Statistics.}
Table~\ref{tab:data_statistics} shows various statistics for the benchmark. 
In total, the dataset includes 1616 tasks and 5M instances.
On average, each instruction is paired with 2.8 positive and 2.4 negative examples. 
The average definition length is 56.6 in words. 

\begin{table}[t]
    \centering
    \small
    \begin{tabular}{lr}
    \hline
    \toprule
    statistic &  \\ 
    \midrule
    \# of tasks & 1616 \\
    \# of task types & 76 \\
    \# of languages & 55 \\
    \# of domains & 33 \\
    \# of non-English tasks & 576 \\
    avg.~definition length (words per task) & 56.6 \\
    avg.~\# of positive examples (per task) & 2.8 \\
    avg.~\# of negative examples (per task) & 2.4 \\ 
    avg.~\# of instances (per task) & 3106.0 \\
    \bottomrule
    \end{tabular}
    \caption{
        Statistics of \nameShort.
    }
    \label{tab:data_statistics}
\end{table}

\section{\modelName{}: Learning to Follow Instructions at Scale}
\label{sec:our-model}
\vskip .0cm \noindent {\bf Defining Generalization to Unseen Tasks.}
\label{subsec:input:output}
Each task $t$ is defined via its natural language instruction $I_t$, and each task has a set of  input/output instances $(X_t, Y_t)$.  
A model $M$ is expected to produce the output $y$, given the input $x$ and the task instruction $I_t$: $M(I_t, x) = y, \;  \mbox{for} \ (x,y)\in (X_t, Y_t)$.
In particular, we would like to evaluate model $M$ on tasks that are \underline{not} observed (i.e., their instances were not used for training 
$M$). 
The only source of signal for learning the task at inference time is in-context instructions $I_t$ that contain a definition and demonstration examples of the task. 

\vskip .1cm \noindent {\bf \modelName{}.}
We introduce {\bf \modelName}, a model that is meta-trained on \nameShort{} for solving tasks given their in-context instructions. 
Previous work has shown the effectiveness of such meta-training in improving model's ability to do in-context learning with either prompts~\cite{zhong2021adapting, sanh2022multitask} or demonstration examples \cite{min2022metaicl}. Because of the large variety of tasks in \nameShort{}, we are able to do this multi-task meta-training at a larger scale than before.
We conduct our experiments and analysis based on the T5 model~\cite{raffel2020exploring}.
Since each instruction $I_t$ consists of multiple elements as described in our instruction schema (\S\ref{sec:dataset}), we map these elements to textual format and append them before the input instance. Fig.~\ref{fig:encoding} in the appendix shows how we encode the full instructions. We study different combinations of these instruction elements in \S\ref{subsec:effects_of_encoding}. By default, we will use our most effective instruction elements (i.e., task definition and two positive examples) unless otherwise specified.
In the same manner, we train the multilingual variant m{\bf \modelName} based on the mT5 model \cite{xue2021mt5}.

\section{Benchmarking Cross-Task Generalization with \nameShort}
\label{sec:empirical}
Here we provide our recommended recipe for 
benchmarking generalization via \nameShort. 

\subsection{Evaluation Setup}
\label{subsec:setup}

\noindent {\bf An Evaluation Split of Unseen Tasks.}
We split the large collection of tasks in \nameShort{} into two subsets: one for evaluation and the other for supervision. 
For evaluation tasks, we fix a manually-selected collection of 12 categories that represent 154 tasks. 
The large variety of tasks in \nameShort{} enables us to choose a diverse set of tasks for evaluation --  such as those at word, sentence, and document levels, covering both classification and generation formats.
Appendix~\ref{sec:eval tasks} lists our evaluation tasks with examples for representative tasks. 
For an efficient evaluation, we sample a maximum of 100 instances for each task, which results in 15,310 testing instances in total.
The remaining tasks are used for 
training models.\footnote{To avoid data leakage, we exclude tasks from the training set if they are sourced from the same dataset as any test task. This results in 757 training tasks for the English track and 1271 training tasks for the cross-lingual track.}

\vskip .1cm \noindent {\bf Divided Tracks for English and X-lignual Tasks.}
\nameShort{} consists of tasks across multiple languages, which  enables evaluating the model's generalization to unseen tasks not only in English but also in other languages. Therefore, we divide our evaluation tasks into two tracks: one for {\bf English cross-task generalization} (119 tasks) and the other for {\bf cross-lingual cross-task generalization} (35 tasks). 
To the best of our knowledge, this is the first study in cross-lingual cross-task generalization (i.e., generalization to unseen tasks in  different languages).
Fig.~\ref{fig:perf_improve_per_task} and Fig.~\ref{fig:perf_improve_per_task_non_en} in the appendix contain the evaluation tasks for each track. 

\vskip .1cm \noindent {\bf Evaluation Metrics.}
\label{subsec:metrics}
Due to the diversity of our tasks and the open-ended generation nature of our formulation,\footnote{Unlike \citet{sanh2022multitask} and \citet{wei2022finetuned}, who evaluate their models on classification tasks
via option ranking 
(i.e., scoring the correct answer(s) higher than other candidate answers), we evaluate our models in an \emph{open-ended} generation setting with no task-specific assumptions. We believe this is a more realistic measure of generalization to unseen tasks.} we adopt \emph{ROUGE-L}~\cite{lin2004rouge} for reporting aggregated performance results. 
This is a soft string overlap metric that can be applied to a wide range of text generation tasks. 
We show that the ranking from this metric correlates well with accuracy for classification tasks in Appendix \ref{sec:metric-correlation}. 
We also conduct a human evaluation in \S\ref{subsec:human_eval}.

\subsection{Baselines and Existing Models}
\label{subsec:baselines}
Here we discuss a variety of baselines and competitive models for our target application. 
See Appendix \ref{sec:implementation} for implementation details. 

\vskip .1cm \noindent {\bf Heuristic baselines.}
We first evaluate the following heuristics to evaluate the possible shortcuts in the data. 
\textbf{Copying Demo Output} copies the output of a random demonstration example. Since we balance the labels for our test tasks, the performance of this baseline will roughly equal a random guess or a majority baseline for classification tasks. \textbf{Copying Instance Input} copies the given instance input. This strategy performs well on tasks where the target output largely overlaps with the input (e.g., question rewriting, grammar error correction).




\vskip .1cm \noindent {\bf Off-the-shelf pretrained language models.}
We evaluate existing 
LMs that are not fine-tuned with instruction-specific data. 
Specifically, we evaluate the 11B-parameter T5~\cite{raffel2020exploring} as a direct counterpart of \modelName{}. Due to the infilling pretraining objective of the original T5 model, it cannot continue text well. Therefore, we evaluate its ``LM-adapted'' version, which is further trained with a language modeling objective~\cite{lester2021power}. 
Additionally, we evaluate GPT-3~\cite{brown2020language}, a $175$B-parameter autoregressive LM that has shown remarkable ability in following demonstrations provided in its prompt.

\vskip .1cm \noindent {\bf Instruction-tuned models.}
In addition to our \modelName{} (\S\ref{sec:our-model}), 
we evaluate existing models that are fine-tuned to follow language instructions. 
In particular, we evaluate InstructGPT~\cite{ouyang2022training} which uses reinforcement learning to incorporate human preferences into a GPT-3 pretrained model, and T0~\cite{sanh2022multitask} which finetunes T5 on a collection of task prompts in \promptsource~\cite{bach2022promptsource}.



\vskip .1cm \noindent {\bf Upper bound estimates.}
We estimate an upper bound on models' generalization to unseen tasks by fine-tuning an \textbf{oracle} model 
on the tasks' labeled instances.
Since this model observes the \emph{hidden} instances of the evaluation tasks, it is, by definition, an estimated upper bound to our generalization-based models. 
Specifically, 
we fine-tune a T5-11B model on the 119 English evaluation tasks, and a mT5-13B model on the 35 non-English tasks, with 1K random training instances per task, without overlap with the evaluation instances.

\begin{table}[t]
\centering
\resizebox{0.99\linewidth}{!}{%
\begin{tabular}{clcc}
\toprule
\multicolumn{2}{c}{\hspace{3cm} Methods $\downarrow$ / Evaluation $\rightarrow$ }                                                                       & En                  & X-lingual              \\ \midrule
\multirow{2}{*}{\makecell{Heuristic\\Baselines}}           & Copying Instance Input               & 14.2                & 5.4                   \\
                                                           & Copying Demo Output                  & 28.5                & 50.3                   \\  \midrule
\multirow{2}{*}{Pretrained LMs}                             & T5-LM (11B)                         & 30.2                & --                     \\
                                                           & GPT3 (175B)                          & 45.0                & 51.3                   \\ \midrule
\multirow{4}{*}{\makecell{Instruction-tuned\\Models}}      & T0 (11B)                             & 32.3                & --                     \\
                                                           & InstructGPT (175B)                 & 52.1                & 52.8                   \\
                                                           & \modelName{} (ours, 11B)            & \bf{62.0}           & --                     \\ 
                                                           & m\modelName{} (ours, 13B)           & 57.1                & \bf{66.1}              \\ \midrule
\color{gray}\multirow{1}{*}{\makecell{Upper-bound (est.)}}      & \color{gray} Supervised Training     & \color{gray}74.3    & \color{gray}94.0       \\
                                                        \bottomrule
\end{tabular}%
}
\caption{
    The overall performance of different methods on unseen tasks in the test set of \nameShort{} (\S\ref{subsec:main_results}). 
    We report ROUGE-L here as our aggregated metric.
    {\bf Models that leverage instructions show stronger generalization to unseen tasks}. In particular, our model that is fine-tuned on a diverse set of tasks outperforms InstructGPT and T0 by a large margin.
}
\label{tab:main_results}
\end{table}

\begin{figure}[ht]
    \centering
    \includegraphics[width=0.95\columnwidth, trim=0cm 0.5cm 0cm 0.4cm]{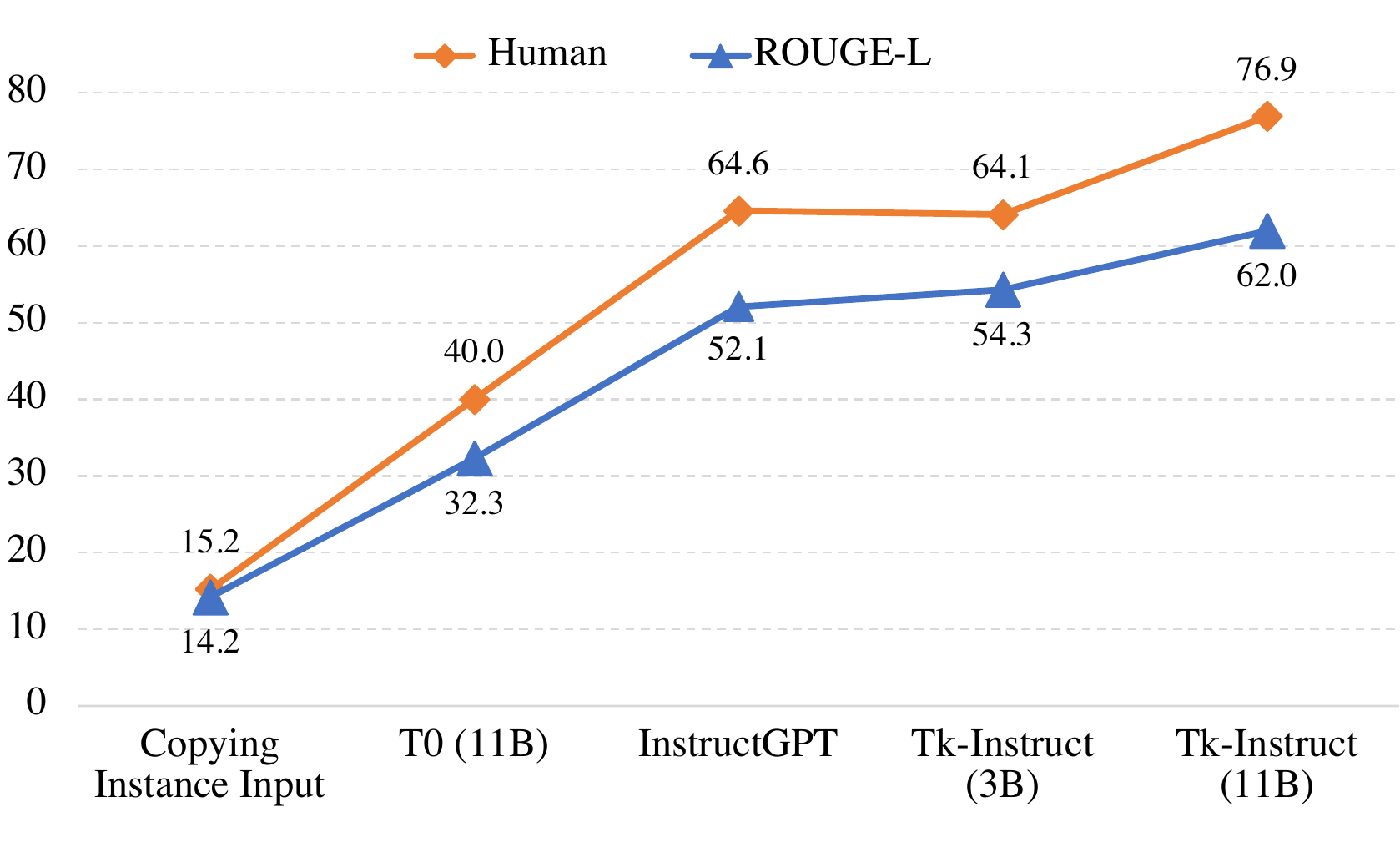}
    \caption{Human evaluation vs. ROUGE-L for several methods (\S\ref{subsec:human_eval}). 
    The trends of these two metrics are highly correlated with a Pearson coefficient of 0.998.}
    \label{fig:human-eval-results}
\end{figure}

\begin{figure*}
    \centering
    \includegraphics[width=\textwidth
    ]{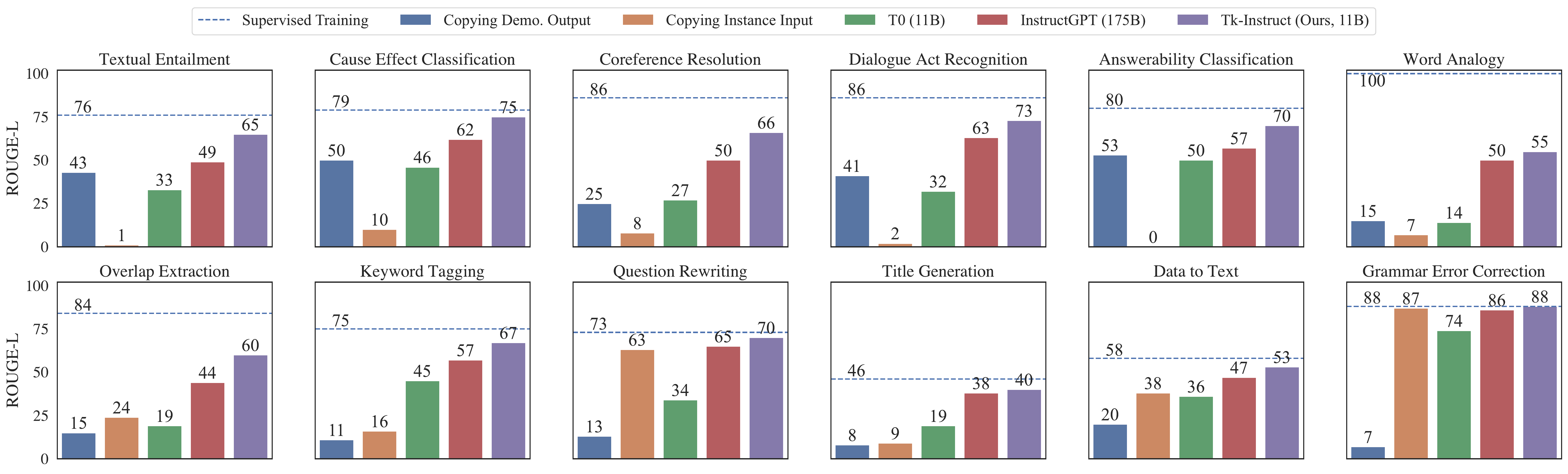}
    \caption{Performance per evaluation task type. 
    \modelName{} consistently performs better than other generalization-based methods on all task types, while there is still a sizable gap compared to supervised training.}
    
    \label{fig:perf-per-category}
\end{figure*}

\section{Experimental Results}

\subsection{Overall Results}
\label{subsec:main_results}
Table~\ref{tab:main_results} summarizes our overall benchmarking results.
We use the same input encoding that contains the most effective instructional elements (task definition and two positive examples without the negative examples and explanations) for all the methods. To better understand models' generalization to different tasks, we also break down the performance according to the task categories in Fig.~\ref{fig:perf-per-category}. We refer the reader to Appendix \ref{sec:per-task performance} for more detailed analysis on each individual task.

\vskip .1cm \noindent {\bf Instruction-tuning enables stronger generalization to unseen tasks.}
Generally instruction-tuned models perform  better compared to their untuned LM counterparts (\modelName{} vs. T5-LM, InstructGPT vs. GPT-3) and heuristic baselines. This indicates models do learn to follow instructions by finetuning on instruction data, and this can generalize to new instructions for unseen tasks. T0 is an exception, which is only slightly better than T5-LM. We suspect this is because the style of prompting in T0's training data is very different from our style of instructions.

\vskip .1cm \noindent {\bf Our \modelName{} outperforms InstructGPT.}
Our \modelName{} and m\modelName{} models, which are trained with a variety of tasks, generalize best to unseen tasks for both English and non-English tasks in all evaluation task categories. InstructGPT also shows a great extent of generalization to our evaluation tasks. However, we want to note it is not clear if InstructGPT's training data overlaps with our evaluation tasks since their data is unavailable. 

\vskip .1cm \noindent {\bf There is a sizable gap for improvement.}
Despite the impressive performance of current models, there is a sizable gap between the generalization of instruction-based models and the supervised training approach,
leaving more room for improvement.

\begin{figure*}[t]
\centering
\begin{subfigure}{.325\textwidth}
    \centering
    \includegraphics[width=\linewidth
    ,trim=0cm 0.55cm 0cm 0.2cm ,clip]{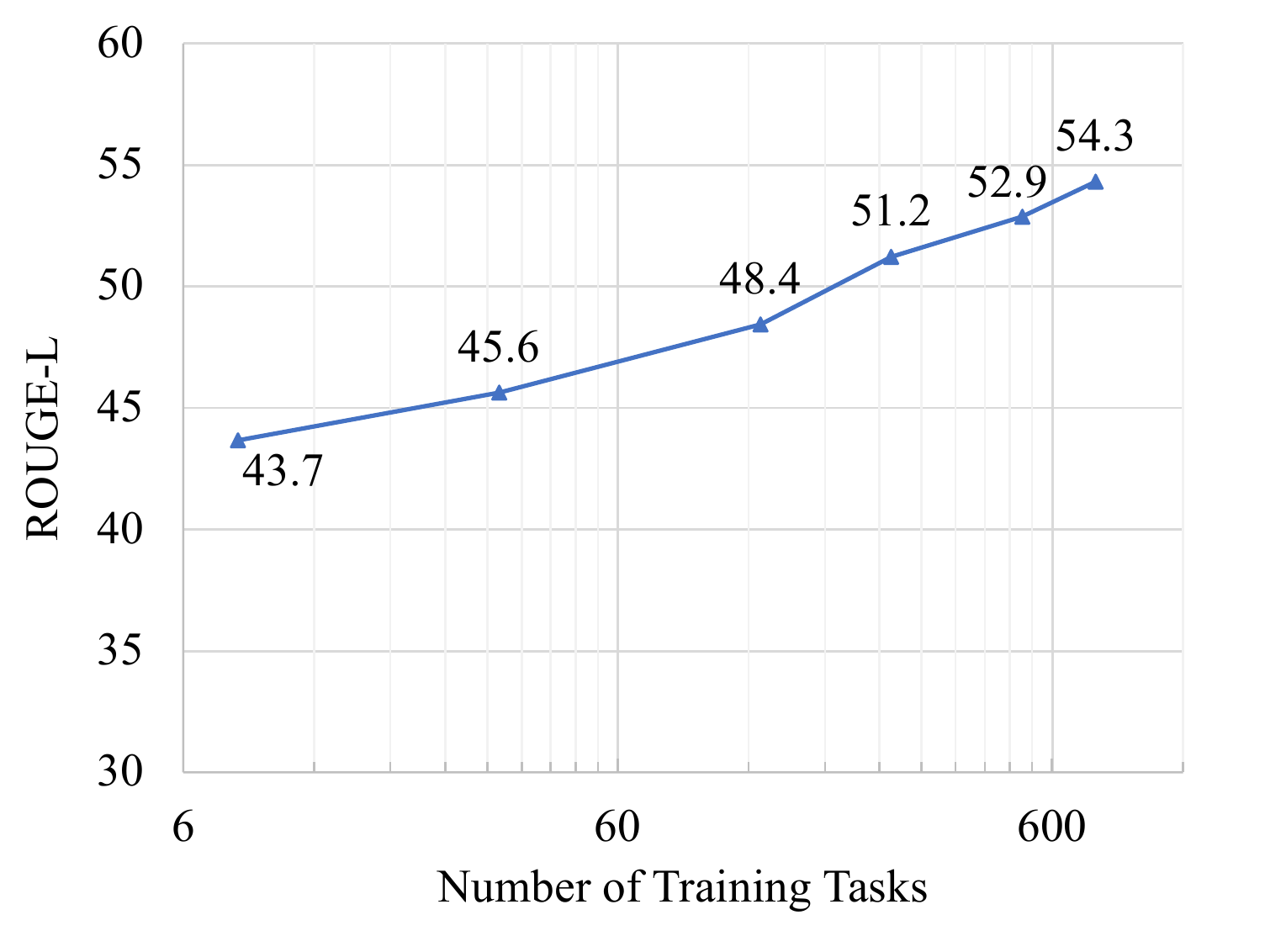} 
    \caption{}
    \label{fig:perf_against_num_tasks}
\end{subfigure}
\begin{subfigure}{.325\textwidth}
    \centering
    \includegraphics[width=\linewidth,trim=0cm 0.55cm 0cm 0.2cm ,clip]{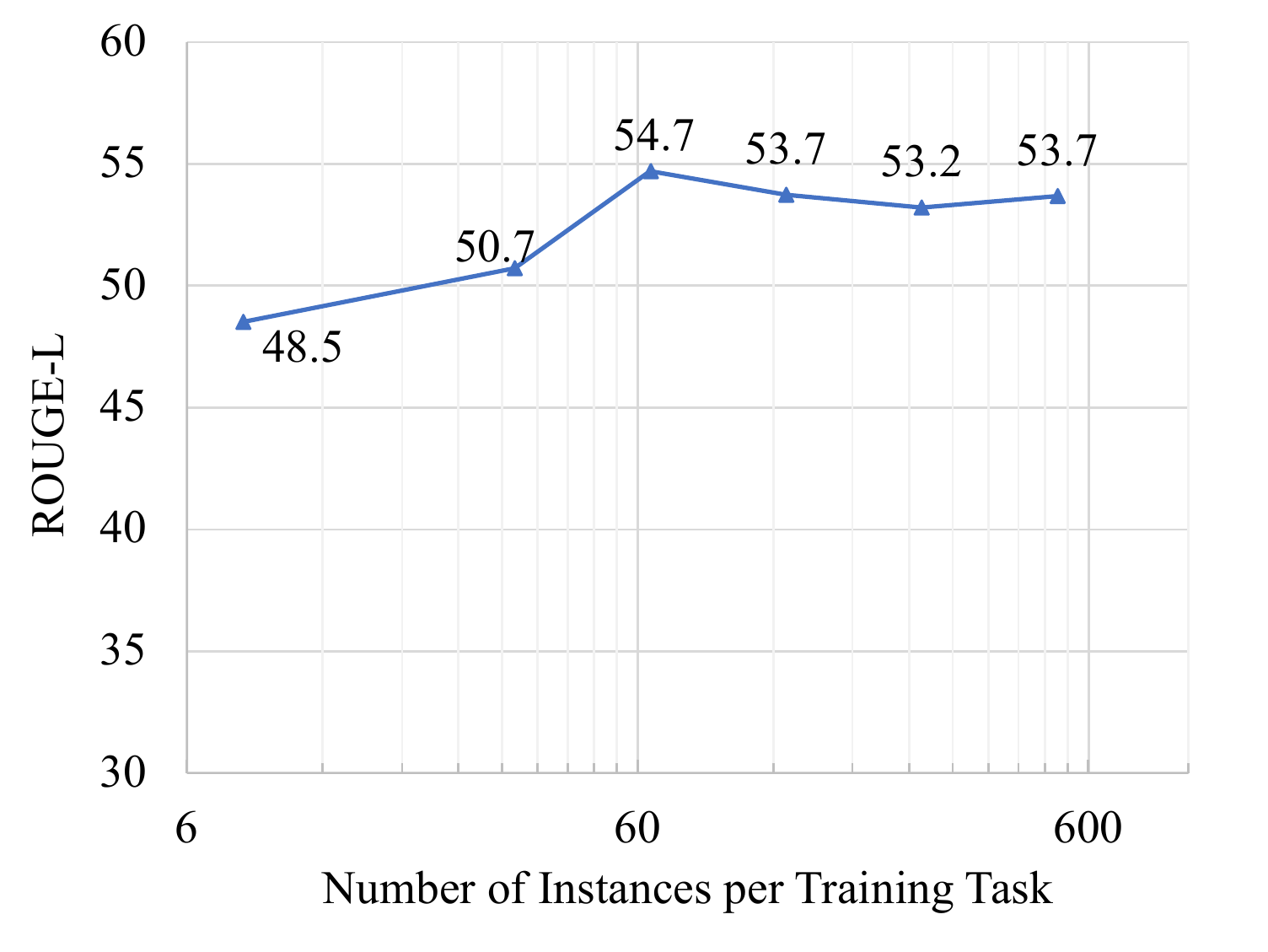}  
    \caption{}
    \label{fig:perf_against_num_instances}
\end{subfigure}
\begin{subfigure}{.325\textwidth}
    \centering
    \includegraphics[width=\linewidth,trim=0cm 0.55cm 0cm 0.2cm ,clip]{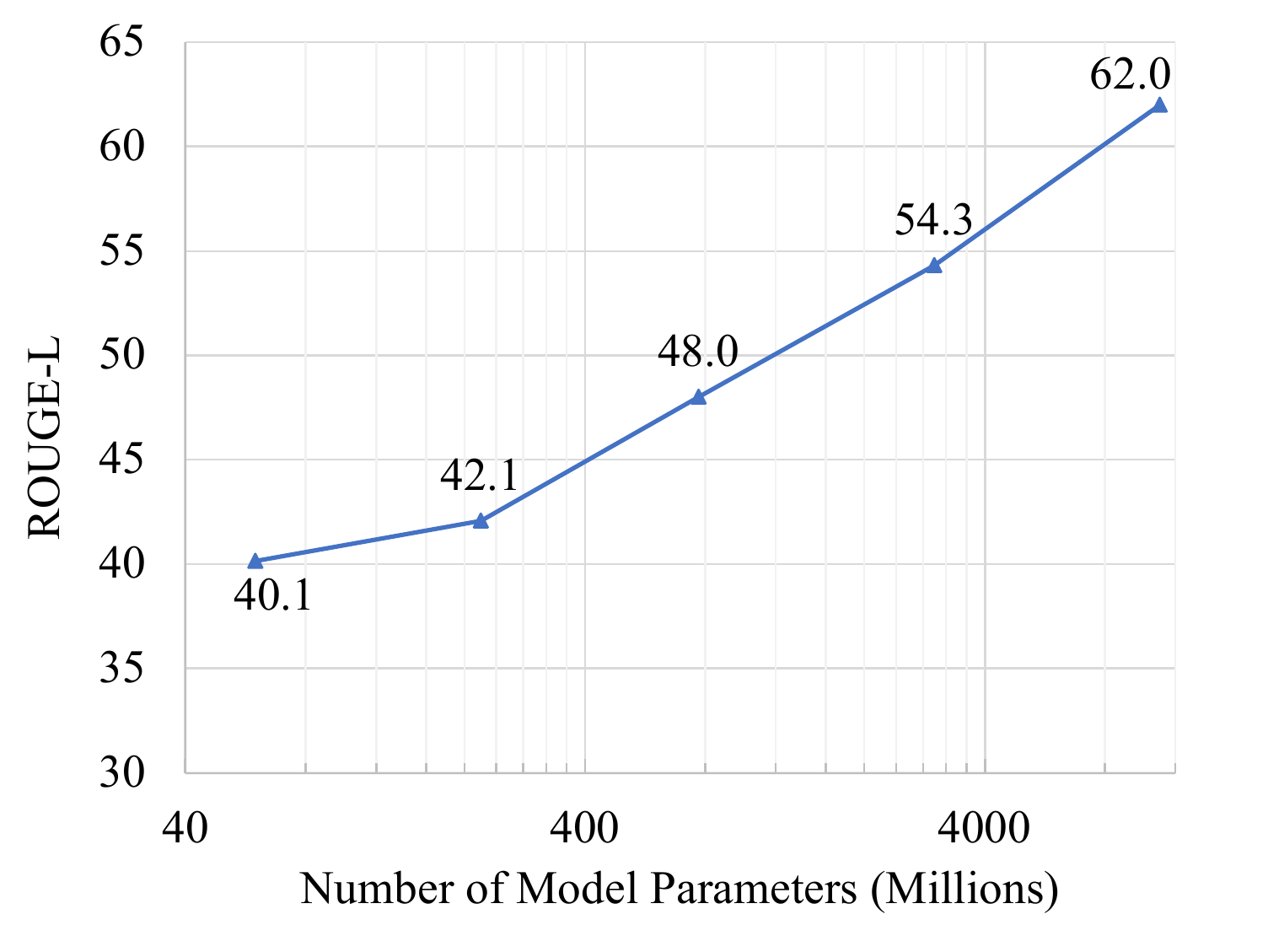}
    \caption{}
    \label{fig:perf_against_model_size}
\end{subfigure}
\caption{
Scaling trends of models performance (\S\ref{subsec:scaling}) as a function of (a) the number of training tasks; (b) the number of instances per training task; (c) model sizes. 
$x$-axes are in log scale.
The {\bf linear growth of  model performance with exponential increase in observed tasks  and model size} is a promising trend. 
Evidently, the performance gain from more instances is limited.  
}
\label{fig:scaling}
\end{figure*}

\subsection{Human Evaluation}
\label{subsec:human_eval}

For language generation tasks, automatic metrics are only an approximation of human judgments;  we conduct a human evaluation to confirm the findings so far.
Specifically, we ask crowdworkers to indicate if they prefer the predicted answer by the model or the ground truth outputs for each instance with ties being allowed (see Appendix~\ref{sec:crowdsourcing:judgements} for details). 
The resulting human evaluation metric indicates how often  model predictions were rated as at least as good as our ground truth labels. 
The theoretical upper bound of this metric is 100\% when the model is rated at least as good as the ground truth for all the instances. 
The results of human evaluation (shown in Fig.~\ref{fig:human-eval-results}) align quite well with our automatic metrics and confirm the human-perceived quality of our models.



\begin{table*}[t!]
    \centering
    \includegraphics[width=0.95\textwidth
    ]{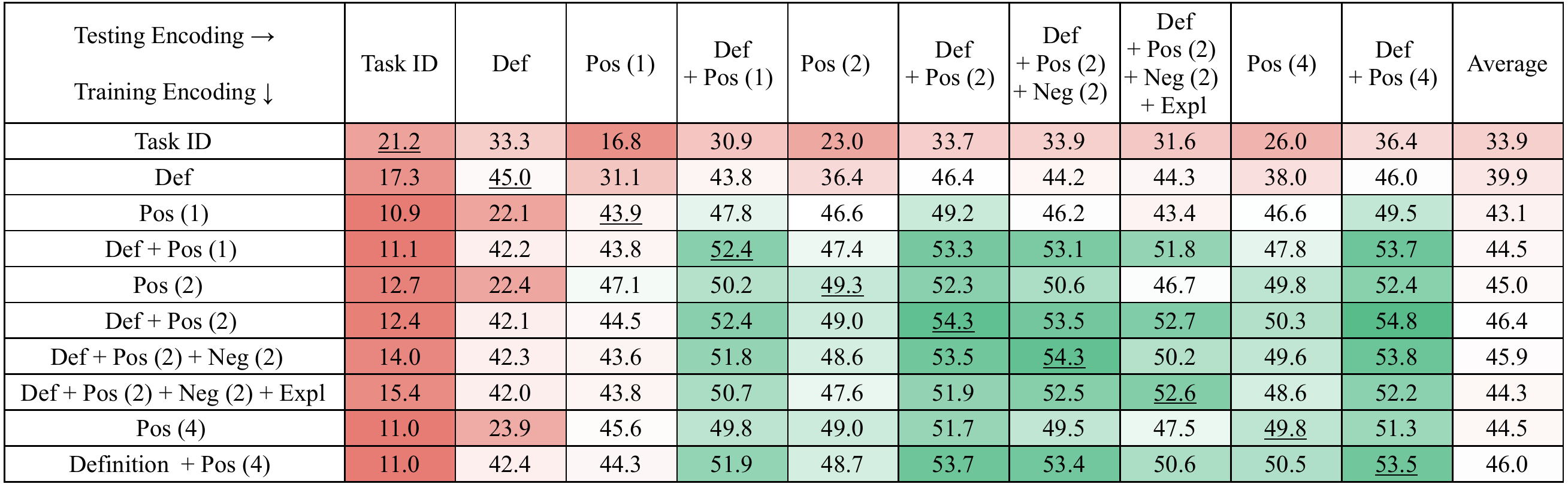}
    \caption{Performance (ROUGE-L) of models trained and evaluated with various encodings. 
    Diagonal numbers (underlined) represent performances of  models trained and evaluated with the same instruction encoding. 
    Each encoding is a combination of the elements in
    the instructions 
    (Fig.~\ref{fig:example}). 
    \textbf{Task ID} is a short string composed of 
    dataset name and task category; \textbf{Def} represents the task definition; \textbf{Pos (k)} represents $k$ positive examples; \textbf{Neg (k)} represents $k$ negative examples; \textbf{Expl} represents explanation.
    These results (a) {\bf show the gains from various instructional elements}, and (b) indicate {\bf surprising reliability of the models to various input encoding}.
    A model trained with definition and positive examples (i.e., the last row) remains robust for different encodings.
    }
    \label{tab:transferring_across_encodings}
\end{table*}

\section{Further Analysis}
\label{sec:analysis}

We conduct further analysis to understand the important factors for models to generalize across tasks. Due to the computational cost, this analysis is done on the English track and using the T5-3B checkpoint, except for the experiments on model sizes. 

\subsection{Scaling Trends of Generalization}
\label{subsec:scaling}


We study \modelName{}'s generalization performance with respect to three scaling factors: the number of training tasks, the number of instances per task, and the model sizes. 
Fig.~\ref{fig:scaling} presents the performance change by scaling each of them. 

\vskip .1cm \noindent {\bf More observed tasks improve the generalization.}
We fine-tune \modelName{} with different numbers of tasks that are randomly sampled from the whole training set (Fig. \ref{fig:perf_against_num_tasks}). The model generalization performance grows log-linearly\footnote{A linear function of an exponential increase of parameters, i.e., growth at a constant multiplicative rate.} as we increase the set of tasks used for training. 
Previous work \cite{mishra2022cross, sanh2022multitask, wei2022finetuned} has made similar observations on a much smaller scale, while we show that this trend holds even with 757 diverse training tasks. 

\vskip .1cm \noindent {\bf A large number of training instances do \underline{not} help generalization.}
We then vary the number of instances per task that are used for finetuning (Fig. \ref{fig:perf_against_num_instances}). 
While the conventional wisdom in supervised learning is that more training instances usually helps~\cite{banko2001scaling,sun2017revisiting,hestness2017deep}, 
in our setup, the model's performance saturates when only 64 instances per task are used for training. A large number of training instances would instead lead to longer training time and risk overfitting to the training tasks.

\vskip .1cm \noindent {\bf Tuning larger models with instructions consistently lead to gains.}
We study the effect of model scaling by initializing \modelName{} from different sizes of pretrained T5 checkpoints, including the small, base, large, xl and xxl sizes (Fig. \ref{fig:perf_against_model_size}). 
We found that increasing the model sizes consistently bring significant improvement (log-linearly with parameter size). 
This finding contradicts the claim in~\citet{xu2022zeroprompt}  that ``model size has little impact on performance with an extremely large number of tasks.''
Combining Fig.~\ref{fig:scaling}(a) and Fig.~\ref{fig:scaling}(c), one can create a correspondence between model size and task size. 
For example, a T5-large model trained with 757 tasks can achieve comparable performance (48.0 ROUGE-L) to the T5-3B model trained with 128 tasks (48.4 ROUGE-L), indicating that increasing the diversity of training tasks is an alternative to scaling model sizes. 

\subsection{Instructing with Different Elements}
We evaluate the performance of \modelName{} under different instructional elements.

\vskip .1cm \noindent {\bf Benefit of different instructional elements.}
\label{subsec:effects_of_encoding}
As shown in Fig.~\ref{fig:example}, \nameShort{} provides multiple elements for instructing a task. 
We train multiple models with different combinations of these elements. 
The \emph{diagonal} cells of  Table~\ref{tab:transferring_across_encodings} show the performance of our models when trained and evaluated on a particular instruction encoding. 
Based on the diagonal numbers, 
including the task definition consistently helps the model to generalize better.
Moreover, combining the task definition with positive demonstration examples yields further improvement.
However, adding more demonstration examples is negligible. Negative examples help a little bit; explanations decrease performance, which is consistent with the observations of \citet{mishra2022cross} and \citet{lampinen2022can} when the model is not large enough. Future work can explore whether more powerful models can benefit from these elements.

\vskip .1cm \noindent {\bf Generalization to different input encodings.}
We further investigate whether a model trained on a particular encoding can generalize to other encodings. 
This can be read from the \emph{non-diagonal} cells of Table~\ref{tab:transferring_across_encodings}. 
The negative result here is that definition-only models cannot generalize to example-only test encodings; and similarly, example-only models cannot generalize to definition-only test encodings.
However, models trained on encodings that contain both definition and examples are surprisingly robust across different encoding variations.

\section{Conclusion}
We construct a large-scale benchmark consisting of a diverse set of NLP tasks and their instructions. 
This benchmark can serve as a rich playground for training or evaluation of models that can generalize to unseen tasks by following instructions. 
Furthermore, we train \modelName{} using this data, and demonstrate its capability to perform unseen tasks to a surprising extent. We provide extensive analysis to understand the important factors for such generalization.
We hope our data and model will facilitate future work towards more general-purpose models.


\section{Limitations}
While the presented data offers a notable variety (e.g., diverse task types), its underlying distributions suffer from skews which should be addressed in future work (see Appendix \ref{sec:distribution}). 
On language diversity, the proposed benchmark is biased toward English. 
On output diversity, the collected tasks are generally still skewed to short responses, 
which   
might reflect 
the distribution of the available tasks in the field. 
This under-representation of the long-tail of tasks
poses a challenge for building general-purpose models in the future.
We hope future work addresses such distributional imbalances. 
Moreover, we see natural extensions of the instruction-following setup here in the context of other modalities such as vision or speech. 

Automatic evaluation of models' performance is another challenge, considering the diverse set of tasks in our benchmark, 
and 
many of them being 
open-ended generation tasks. 
We use ROUGE-L as an aggregated metric in this paper and find it as a good proxy for the \emph{overall} performance of the models, aligning well with human evaluation. 
However, there are specific tasks for which ROUGE-L might not serve as an effective proxy of quality (such as rewriting tasks or error correction tasks where copying the input can result in a high ROUGE-L score). 
We hope these issues will be addressed with the development of more powerful evaluation metrics for text generation.


In terms of computing power, we have experimented with models that were accessible to us and have made the resulting models publicly available.
We also acknowledge that there are larger models that we were not able to train due to the limitations of our computational budget. 


\section*{Acknowledgments}
We thank the anonymous reviewers, our colleagues from AI2 and UWNLP, especially Matthew Peters for his encouraging conversations that motivated this project. 
We also thank the student contributors of Arizona State University's CSE 576 ``Topics in NLP'' course and all other contributors to our data repository.
All experiments were run on AI2's Beaker GPU clusters and Google's research TPUs. 
This work was supported in part by ONR MURI N00014-18-1-2670, ONR N00014-18-1-2826, and DARPA MCS N66001-19-2-4031 grants.


\bibliography{ref}

\providecommand{\CNFX}[1]{{\em{\textrm{(#1)}}}}
\begin{thebibliography}{54}
\expandafter\ifx\csname natexlab\endcsname\relax\def\natexlab#1{#1}\fi

\bibitem[{Aribandi et~al.(2022)Aribandi, Tay, Schuster, Rao, Zheng, Mehta,
  Zhuang, Tran, Bahri, Ni et~al.}]{aribandi2022ext5}
Vamsi Aribandi, Yi~Tay, Tal Schuster, Jinfeng Rao, Huaixiu~Steven Zheng,
  Sanket~Vaibhav Mehta, Honglei Zhuang, Vinh~Q Tran, Dara Bahri, Jianmo Ni,
  et~al. 2022.
\newblock \href {https://arxiv.org/abs/2111.10952} {{ExT5: Towards Extreme
  Multi-Task Scaling for Transfer Learning}}.
\newblock In \emph{International Conference on Learning Representations
  \CNFX{ICLR}}.

\bibitem[{Bach et~al.(2022)Bach, Sanh, Yong, Webson, Raffel, Nayak, Sharma,
  Kim, Bari, Fevry et~al.}]{bach2022promptsource}
Stephen~H Bach, Victor Sanh, Zheng-Xin Yong, Albert Webson, Colin Raffel,
  Nihal~V Nayak, Abheesht Sharma, Taewoon Kim, M~Saiful Bari, Thibault Fevry,
  et~al. 2022.
\newblock \href {https://arxiv.org/abs/2202.01279} {{PromptSource: An
  Integrated Development Environment and Repository for Natural Language
  Prompts}}.
\newblock In \emph{Annual Meeting of the Association for Computational
  Linguistics \CNFX{ACL} - System Demonstrations}.

\bibitem[{Banko and Brill(2001)}]{banko2001scaling}
Michele Banko and Eric Brill. 2001.
\newblock \href {https://aclanthology.org/P01-1005/} {{Scaling to Very Very
  Large Corpora for Natural Language Disambiguation}}.
\newblock In \emph{Annual Meeting of the Association for Computational
  Linguistics \CNFX{ACL}}.

\bibitem[{Bartolo et~al.(2020)Bartolo, Roberts, Welbl, Riedel, and
  Stenetorp}]{bartolo2020beat}
Max Bartolo, Alastair Roberts, Johannes Welbl, Sebastian Riedel, and Pontus
  Stenetorp. 2020.
\newblock \href {https://arxiv.org/abs/2002.00293} {Beat the ai: Investigating
  adversarial human annotation for reading comprehension}.
\newblock \emph{Transactions of the Association for Computational Linguistics
  \CNFX{TACL}}, 8:662--678.

\bibitem[{Bastan et~al.(2020)Bastan, Koupaee, Son, Sicoli, and
  Balasubramanian}]{bastan2020author}
Mohaddeseh Bastan, Mahnaz Koupaee, Youngseo Son, Richard Sicoli, and Niranjan
  Balasubramanian. 2020.
\newblock \href {https://arxiv.org/abs/2011.06128} {Author's sentiment
  prediction}.
\newblock In \emph{International Conference on Computational Linguistics
  \CNFX{COLING}}.

\bibitem[{Bentivogli et~al.(2008)Bentivogli, Clark, Dagan, and
  Giampiccolo}]{bentivogli2008the}
Luisa Bentivogli, Peter Clark, Ido Dagan, and Danilo Giampiccolo. 2008.
\newblock \href
  {https://www.semanticscholar.org/paper/The-Sixth-PASCAL-Recognizing-Textual-Entailment-Bentivogli-Clark/db8885a0037fe47d973ade79d696586453710233}
  {The sixth pascal recognizing textual entailment challenge}.
\newblock In \emph{Text Analysis Conference \CNFX{TAC}}.

\bibitem[{Bragg et~al.(2021)Bragg, Cohan, Lo, and Beltagy}]{bragg2021flex}
Jonathan Bragg, Arman Cohan, Kyle Lo, and Iz~Beltagy. 2021.
\newblock \href {https://arxiv.org/abs/2107.07170} {Flex: Unifying evaluation
  for few-shot nlp}.
\newblock In \emph{Advances in Neural Information Processing Systems
  \CNFX{NeurIPS}}.

\bibitem[{Brown et~al.(2020)Brown, Mann, Ryder, Subbiah, Kaplan, Dhariwal,
  Neelakantan, Shyam, Sastry, Askell, Agarwal, and et~al.}]{brown2020language}
Tom~B. Brown, Benjamin Mann, Nick Ryder, Melanie Subbiah, Jared Kaplan,
  Prafulla Dhariwal, Arvind Neelakantan, Pranav Shyam, Girish Sastry, Amanda
  Askell, Sandhini Agarwal, and et~al. 2020.
\newblock \href {http://arxiv.org/abs/2005.14165} {{Language models are
  few-shot learners}}.
\newblock In \emph{Advances in Neural Information Processing Systems
  \CNFX{NeurIPS}}.

\bibitem[{Caruana(1997)}]{caruana1997multitask}
Rich Caruana. 1997.
\newblock \href {https://link.springer.com/article/10.1023/A:1007379606734}
  {Multitask learning}.
\newblock \emph{Machine learning}, 28(1):41--75.

\bibitem[{Chawla et~al.(2021)Chawla, Ramirez, Clever, Lucas, May, and
  Gratch}]{chawla2021casino}
Kushal Chawla, Jaysa Ramirez, Rene Clever, Gale Lucas, Jonathan May, and
  Jonathan Gratch. 2021.
\newblock \href {https://arxiv.org/abs/2103.15721} {{CaSiNo: A Corpus of
  Campsite Negotiation Dialogues for Automatic Negotiation Systems}}.
\newblock In \emph{Conference of the North American Chapter of the Association
  for Computational Linguistics \CNFX{NAACL}}.

\bibitem[{Collobert and Weston(2008)}]{collobert2008unified}
Ronan Collobert and Jason Weston. 2008.
\newblock \href {http://machinelearning.org/archive/icml2008/papers/391.pdf} {A
  unified architecture for natural language processing: Deep neural networks
  with multitask learning}.
\newblock In \emph{International Conference on Machine Learning \CNFX{ICML}}.

\bibitem[{Dagan et~al.(2005)Dagan, Glickman, and Magnini}]{dagan2005pascal}
Ido Dagan, Oren Glickman, and Bernardo Magnini. 2005.
\newblock \href {https://link.springer.com/chapter/10.1007/11736790\_9} {The
  pascal recognising textual entailment challenge}.
\newblock In \emph{Machine Learning Challenges Workshop}.

\bibitem[{Efrat and Levy(2020)}]{efrat2020turking}
Avia Efrat and Omer Levy. 2020.
\newblock \href {https://arxiv.org/abs/2010.11982} {{The Turking Test: Can
  Language Models Understand Instructions?}}
\newblock \emph{arXiv preprint arXiv:2010.11982}.

\bibitem[{Fulda et~al.(2017)Fulda, Tibbetts, Brown, and
  Wingate}]{fulda2017harvesting}
Nancy Fulda, Nathan Tibbetts, Zachary Brown, and David Wingate. 2017.
\newblock \href {https://proceedings.mlr.press/v78/fulda17a.html} {Harvesting
  common-sense navigational knowledge for robotics from uncurated text
  corpora}.
\newblock In \emph{Conference on Robot Learning \CNFX{IJCAI}}.

\bibitem[{Gupta et~al.(2021)Gupta, Xu, Upadhyay, Yang, and
  Faruqui}]{gupta2021disfl}
Aditya Gupta, Jiacheng Xu, Shyam Upadhyay, Diyi Yang, and Manaal Faruqui. 2021.
\newblock \href {https://aclanthology.org/2021.findings-acl.293.pdf} {Disfl-qa:
  A benchmark dataset for understanding disfluencies in question answering}.
\newblock In \emph{Annual Meeting of the Association for Computational
  Linguistics \CNFX{ACL} - Findings}.

\bibitem[{Hashimoto et~al.(2017)Hashimoto, Xiong, Tsuruoka, and
  Socher}]{hashimoto2017a}
Kazuma Hashimoto, Caiming Xiong, Yoshimasa Tsuruoka, and Richard Socher. 2017.
\newblock \href {https://arxiv.org/abs/1611.01587} {A joint many-task model:
  Growing a neural network for multiple nlp tasks}.
\newblock In \emph{Conference on Empirical Methods in Natural Language
  Processing \CNFX{EMNLP}}.

\bibitem[{Hersh et~al.(1994)Hersh, Buckley, Leone, and
  Hickam}]{hersh1994ohsumed}
William Hersh, Chris Buckley, TJ~Leone, and David Hickam. 1994.
\newblock \href
  {https://link.springer.com/chapter/10.1007/978-1-4471-2099-5\_20} {Ohsumed:
  An interactive retrieval evaluation and new large test collection for
  research}.
\newblock In \emph{Conference of the Association for Computing Machinery
  Special Interest Group in Information Retrieval \CNFX{SIGIR}}.

\bibitem[{Hestness et~al.(2017)Hestness, Narang, Ardalani, Diamos, Jun,
  Kianinejad, Patwary, Ali, Yang, and Zhou}]{hestness2017deep}
Joel Hestness, Sharan Narang, Newsha Ardalani, Gregory Diamos, Heewoo Jun,
  Hassan Kianinejad, Md~Patwary, Mostofa Ali, Yang Yang, and Yanqi Zhou. 2017.
\newblock \href {https://arxiv.org/abs/1712.00409} {{Deep Learning Scaling is
  Predictable, Empirically}}.
\newblock \emph{arXiv preprint arXiv:1712.00409}.

\bibitem[{Khashabi et~al.(2020)Khashabi, Min, Khot, Sabharwal, Tafjord, Clark,
  and Hajishirzi}]{khashabi2020unifiedqa}
Daniel Khashabi, Sewon Min, Tushar Khot, Ashish Sabharwal, Oyvind Tafjord,
  Peter Clark, and Hannaneh Hajishirzi. 2020.
\newblock \href {https://arxiv.org/abs/2005.00700} {{UnifiedQA: Crossing Format
  Boundaries With a Single QA System}}.
\newblock In \emph{Conference on Empirical Methods in Natural Language
  Processing \CNFX{EMNLP} - Findings}.

\bibitem[{Kim et~al.(2020)Kim, So, Han, and Lee}]{kim2020natural}
Hyeonji Kim, Byeong-Hoon So, Wook-Shin Han, and Hongrae Lee. 2020.
\newblock \href {https://www.vldb.org/pvldb/vol13/p1737-kim.pdf} {Natural
  language to sql: Where are we today?}
\newblock \emph{Proceedings of the VLDB Endowment}, 13(10):1737--1750.

\bibitem[{Lampinen et~al.(2022)Lampinen, Dasgupta, Chan, Matthewson, Tessler,
  Creswell, McClelland, Wang, and Hill}]{lampinen2022can}
Andrew~K Lampinen, Ishita Dasgupta, Stephanie~CY Chan, Kory Matthewson,
  Michael~Henry Tessler, Antonia Creswell, James~L McClelland, Jane~X Wang, and
  Felix Hill. 2022.
\newblock \href {https://arxiv.org/abs/2204.02329} {{Can Language Models Learn
  from Explanations in Context?}}
\newblock \emph{arXiv preprint arXiv:2204.02329}.

\bibitem[{Lebanoff et~al.(2020)Lebanoff, Muchovej, Dernoncourt, Kim, Wang,
  Chang, and Liu}]{lebanoff2020understanding}
Logan Lebanoff, John Muchovej, Franck Dernoncourt, Doo~Soon Kim, Lidan Wang,
  Walter Chang, and Fei Liu. 2020.
\newblock \href {https://aclanthology.org/2020.acl-srw.26/} {Understanding
  points of correspondence between sentences for abstractive summarization}.
\newblock In \emph{Annual Meeting of the Association for Computational
  Linguistics \CNFX{ACL} - Student Research Workshop}.

\bibitem[{Lester et~al.(2021)Lester, Al-Rfou, and Constant}]{lester2021power}
Brian Lester, Rami Al-Rfou, and Noah Constant. 2021.
\newblock \href {https://arxiv.org/abs/2104.08691} {The power of scale for
  parameter-efficient prompt tuning}.
\newblock In \emph{Conference on Empirical Methods in Natural Language
  Processing \CNFX{EMNLP}}.

\bibitem[{Lin(2004)}]{lin2004rouge}
Chin-Yew Lin. 2004.
\newblock \href {https://aclanthology.org/W04-1013/} {{ROUGE: A Package for
  Automatic Evaluation of Summaries}}.
\newblock In \emph{ACL Workshop on Text Summarization Branches Out}.

\bibitem[{McCann et~al.(2018)McCann, Keskar, Xiong, and
  Socher}]{mccann2018natural}
Bryan McCann, Nitish~Shirish Keskar, Caiming Xiong, and Richard Socher. 2018.
\newblock \href {https://arxiv.org/abs/1806.08730} {The natural language
  decathlon: Multitask learning as question answering}.
\newblock \emph{arXiv preprint arXiv:1806.08730}.

\bibitem[{Min et~al.(2022{\natexlab{a}})Min, Lewis, Zettlemoyer, and
  Hajishirzi}]{min2022metaicl}
Sewon Min, Mike Lewis, Luke Zettlemoyer, and Hannaneh Hajishirzi.
  2022{\natexlab{a}}.
\newblock \href {https://arxiv.org/abs/2110.15943} {Meta{ICL}: Learning to
  learn in context}.
\newblock In \emph{NAACL-HLT}.

\bibitem[{Min et~al.(2022{\natexlab{b}})Min, Chaplot, Ravikumar, Bisk, and
  Salakhutdinov}]{min2022film}
So~Yeon Min, Devendra~Singh Chaplot, Pradeep Ravikumar, Yonatan Bisk, and
  Ruslan Salakhutdinov. 2022{\natexlab{b}}.
\newblock \href {https://arxiv.org/abs/2110.07342} {{FILM: Following
  Instructions in Language with Modular Methods}}.
\newblock In \emph{International Conference on Learning Representations
  \CNFX{ICLR}}.

\bibitem[{Mishra et~al.(2022{\natexlab{a}})Mishra, Khashabi, Baral, Choi, and
  Hajishirzi}]{mishra2022reframing}
Swaroop Mishra, Daniel Khashabi, Chitta Baral, Yejin Choi, and Hannaneh
  Hajishirzi. 2022{\natexlab{a}}.
\newblock \href {https://arxiv.org/abs/2109.07830} {Reframing instructional
  prompts to gptk's language}.
\newblock In \emph{Annual Meeting of the Association for Computational
  Linguistics \CNFX{ACL} - Findings}.

\bibitem[{Mishra et~al.(2022{\natexlab{b}})Mishra, Khashabi, Baral, and
  Hajishirzi}]{mishra2022cross}
Swaroop Mishra, Daniel Khashabi, Chitta Baral, and Hannaneh Hajishirzi.
  2022{\natexlab{b}}.
\newblock \href {https://arxiv.org/abs/2104.08773} {{Cross-Task Generalization
  via Natural Language Crowdsourcing Instructions}}.
\newblock In \emph{Annual Meeting of the Association for Computational
  Linguistics \CNFX{ACL}}.

\bibitem[{Napoles et~al.(2017)Napoles, Sakaguchi, and
  Tetreault}]{napoles2017jfleg}
Courtney Napoles, Keisuke Sakaguchi, and Joel Tetreault. 2017.
\newblock \href {https://aclanthology.org/E17-2037/} {Jfleg: A fluency corpus
  and benchmark for grammatical error correction}.
\newblock In \emph{Conference of the European Chapter of the Association for
  Computational Linguistics \CNFX{EACL}}.

\bibitem[{Novikova et~al.(2017)Novikova, Du{\v{s}}ek, and
  Rieser}]{novikova2017e2e}
Jekaterina Novikova, Ond{\v{r}}ej Du{\v{s}}ek, and Verena Rieser. 2017.
\newblock \href {https://aclanthology.org/W17-5525/} {The e2e dataset: New
  challenges for end-to-end generation}.
\newblock In \emph{Annual SIGdial Meeting on Discourse and Dialogue}.

\bibitem[{Ouyang et~al.(2022)Ouyang, Wu, Jiang, Almeida, Wainwright, Mishkin,
  Zhang, Agarwal, Slama, Ray et~al.}]{ouyang2022training}
Long Ouyang, Jeff Wu, Xu~Jiang, Diogo Almeida, Carroll~L Wainwright, Pamela
  Mishkin, Chong Zhang, Sandhini Agarwal, Katarina Slama, Alex Ray, et~al.
  2022.
\newblock \href {https://arxiv.org/abs/2203.02155} {{Training Language Models
  to Follow Instructions with Human Feedback}}.
\newblock \emph{arXiv preprint arXiv:2203.02155}.

\bibitem[{Parmar et~al.(2022)Parmar, Mishra, Purohit, Luo, Mohammad, and
  Baral}]{parmar-etal-2022-boxbart}
Mihir Parmar, Swaroop Mishra, Mirali Purohit, Man Luo, Murad Mohammad, and
  Chitta Baral. 2022.
\newblock \href {https://doi.org/10.18653/v1/2022.findings-naacl.10}
  {In-{B}o{XBART}: Get instructions into biomedical multi-task learning}.
\newblock In \emph{Findings of the Association for Computational Linguistics:
  NAACL 2022}, pages 112--128, Seattle, United States. Association for
  Computational Linguistics.

\bibitem[{Pavlick et~al.(2014)Pavlick, Post, Irvine, Kachaev, and
  Callison-Burch}]{pavlick2014language}
Ellie Pavlick, Matt Post, Ann Irvine, Dmitry Kachaev, and Chris Callison-Burch.
  2014.
\newblock \href {https://aclanthology.org/Q14-1007/} {{The Language
  Demographics of Amazon Mechanical Turk}}.
\newblock \emph{Transactions of the Association for Computational Linguistics
  \CNFX{TACL}}.

\bibitem[{Raffel et~al.(2020)Raffel, Shazeer, Roberts, Lee, Narang, Matena,
  Zhou, Li, and Liu}]{raffel2020exploring}
Colin Raffel, Noam Shazeer, Adam Roberts, Katherine Lee, Sharan Narang, Michael
  Matena, Yanqi Zhou, Wei Li, and Peter~J Liu. 2020.
\newblock \href {https://arxiv.org/abs/1910.10683} {Exploring the limits of
  transfer learning with a unified text-to-text transformer}.
\newblock \emph{Journal of Machine Learning Research \CNFX{JMLR}}.

\bibitem[{Roemmele et~al.(2011)Roemmele, Bejan, and
  Gordon}]{roemmele2011choice}
Melissa Roemmele, Cosmin~Adrian Bejan, and Andrew~S Gordon. 2011.
\newblock \href {http://commonsensereasoning.org/2011/papers/Roemmele.pdf}
  {Choice of plausible alternatives: An evaluation of commonsense causal
  reasoning.}
\newblock In \emph{AAAI spring symposium: logical formalizations of commonsense
  reasoning}.

\bibitem[{Sakaguchi et~al.(2020)Sakaguchi, Bras, Bhagavatula, and
  Choi}]{sakaguchi2020winogrande}
Keisuke Sakaguchi, Ronan~Le Bras, Chandra Bhagavatula, and Yejin Choi. 2020.
\newblock \href {https://arxiv.org/abs/1907.10641} {{WINOGRANDE:} an
  adversarial winograd schema challenge at scale}.
\newblock In \emph{Conference on Artificial Intelligence \CNFX{AAAI}}.

\bibitem[{Sanh et~al.(2022)Sanh, Webson, Raffel, Bach, Sutawika, Alyafeai,
  Chaffin, Stiegler, Raja, Dey, Bari, Xu, Thakker, Sharma, Szczechla, Kim,
  Chhablani, Nayak, Datta, Chang, Jiang, Wang, Manica, Shen, Yong, Pandey,
  Bawden, Wang, Neeraj, Rozen, Sharma, Santilli, Fevry, Fries, Teehan, Scao,
  Biderman, Gao, Wolf, and Rush}]{sanh2022multitask}
Victor Sanh, Albert Webson, Colin Raffel, Stephen Bach, Lintang Sutawika, Zaid
  Alyafeai, Antoine Chaffin, Arnaud Stiegler, Arun Raja, Manan Dey, M~Saiful
  Bari, Canwen Xu, Urmish Thakker, Shanya~Sharma Sharma, Eliza Szczechla,
  Taewoon Kim, Gunjan Chhablani, Nihal Nayak, Debajyoti Datta, Jonathan Chang,
  Mike Tian-Jian Jiang, Han Wang, Matteo Manica, Sheng Shen, Zheng~Xin Yong,
  Harshit Pandey, Rachel Bawden, Thomas Wang, Trishala Neeraj, Jos Rozen,
  Abheesht Sharma, Andrea Santilli, Thibault Fevry, Jason~Alan Fries, Ryan
  Teehan, Teven~Le Scao, Stella Biderman, Leo Gao, Thomas Wolf, and Alexander~M
  Rush. 2022.
\newblock \href {https://arxiv.org/abs/2110.08207} {{Multitask Prompted
  Training Enables Zero-Shot Task Generalization}}.
\newblock In \emph{International Conference on Learning Representations
  \CNFX{ICLR}}.

\bibitem[{Shalyminov et~al.(2020)Shalyminov, Sordoni, Atkinson, and
  Schulz}]{shalyminov2020fast}
Igor Shalyminov, Alessandro Sordoni, Adam Atkinson, and Hannes Schulz. 2020.
\newblock \href {https://ieeexplore.ieee.org/document/9053599} {Fast domain
  adaptation for goal-oriented dialogue using a hybrid generative-retrieval
  transformer}.
\newblock In \emph{IEEE International Conference on Acoustics, Speech and
  Signal Processing (ICASSP)}.

\bibitem[{Shridhar et~al.(2020)Shridhar, Thomason, Gordon, Bisk, Han, Mottaghi,
  Zettlemoyer, and Fox}]{shridhar2020alfred}
Mohit Shridhar, Jesse Thomason, Daniel Gordon, Yonatan Bisk, Winson Han,
  Roozbeh Mottaghi, Luke Zettlemoyer, and Dieter Fox. 2020.
\newblock \href {https://arxiv.org/abs/1912.01734} {{ALFRED: A Benchmark for
  Interpreting Grounded Instructions for Everyday Tasks}}.
\newblock In \emph{IEEE Conference on Computer Vision and Pattern Recognition
  \CNFX{CVPR}}.

\bibitem[{Srivastava et~al.(2022)Srivastava, Rastogi, Rao, Shoeb, Abid, Fisch,
  Brown, Santoro, Gupta, Garriga-Alonso et~al.}]{srivastava2022beyond}
Aarohi Srivastava, Abhinav Rastogi, Abhishek Rao, Abu Awal~Md Shoeb, Abubakar
  Abid, Adam Fisch, Adam~R. Brown, Adam Santoro, Aditya Gupta, Adri\`{a}
  Garriga-Alonso, et~al. 2022.
\newblock \href {https://arxiv.org/abs/2206.04615} {Beyond the imitation game:
  Quantifying and extrapolating the capabilities of language models}.
\newblock \emph{arXiv preprint arXiv:2206.04615}.

\bibitem[{Stepputtis et~al.(2020)Stepputtis, Campbell, Phielipp, Lee, Baral,
  and Ben~Amor}]{stepputtis2020language}
Simon Stepputtis, Joseph Campbell, Mariano Phielipp, Stefan Lee, Chitta Baral,
  and Heni Ben~Amor. 2020.
\newblock \href {https://arxiv.org/abs/2010.12083} {{Language-Conditioned
  Imitation Learning for Robot Manipulation Tasks}}.
\newblock In \emph{Advances in Neural Information Processing Systems
  \CNFX{NeurIPS}}.

\bibitem[{Sun et~al.(2017)Sun, Shrivastava, Singh, and
  Gupta}]{sun2017revisiting}
Chen Sun, Abhinav Shrivastava, Saurabh Singh, and Abhinav Gupta. 2017.
\newblock \href {https://arxiv.org/abs/1707.02968} {{Revisiting Unreasonable
  Effectiveness of Data in Deep Learning Era}}.
\newblock In \emph{International Conference on Computer Vision \CNFX{ICCV}}.

\bibitem[{Triantafillou et~al.(2019)Triantafillou, Zhu, Dumoulin, Lamblin,
  Evci, Xu, Goroshin, Gelada, Swersky, Manzagol et~al.}]{triantafillou2019meta}
Eleni Triantafillou, Tyler Zhu, Vincent Dumoulin, Pascal Lamblin, Utku Evci,
  Kelvin Xu, Ross Goroshin, Carles Gelada, Kevin Swersky, Pierre-Antoine
  Manzagol, et~al. 2019.
\newblock \href {https://arxiv.org/abs/1903.03096} {{Meta-Dataset: A Dataset of
  Datasets for Learning to Learn from Few Examples}}.
\newblock In \emph{International Conference on Learning Representations
  \CNFX{ICLR}}.

\bibitem[{Wei et~al.(2022)Wei, Bosma, Zhao, Guu, Yu, Lester, Du, Dai, and
  Le}]{wei2022finetuned}
Jason Wei, Maarten Bosma, Vincent Zhao, Kelvin Guu, Adams~Wei Yu, Brian Lester,
  Nan Du, Andrew~M. Dai, and Quoc~V Le. 2022.
\newblock \href {https://arxiv.org/abs/2109.01652} {{Finetuned Language Models
  are Zero-Shot Learners}}.
\newblock In \emph{International Conference on Learning Representations
  \CNFX{ICLR}}.

\bibitem[{Weir et~al.(2022)Weir, Yuan, C{\^o}t{\'e}, Hausknecht, Laroche,
  Momennejad, Van~Seijen, and Van~Durme}]{weir2022one}
Nathaniel Weir, Xingdi Yuan, Marc-Alexandre C{\^o}t{\'e}, Matthew Hausknecht,
  Romain Laroche, Ida Momennejad, Harm Van~Seijen, and Benjamin Van~Durme.
  2022.
\newblock \href {https://arxiv.org/abs/2203.04806} {{One-Shot Learning from a
  Demonstration with Hierarchical Latent Language}}.
\newblock \emph{arXiv preprint arXiv:2203.04806}.

\bibitem[{Weller et~al.(2020)Weller, Lourie, Gardner, and
  Peters}]{weller2020learning}
Orion Weller, Nicholas Lourie, Matt Gardner, and Matthew Peters. 2020.
\newblock \href {https://aclanthology.org/2020.emnlp-main.105/} {{Learning from
  Task Descriptions}}.
\newblock In \emph{Conference on Empirical Methods in Natural Language
  Processing \CNFX{EMNLP}}.

\bibitem[{Wolf et~al.(2020)Wolf, Debut, Sanh, Chaumond, Delangue, Moi, Cistac,
  Rault, Louf, Funtowicz, Davison, Shleifer, von Platen, Ma, Jernite, Plu, Xu,
  Scao, Gugger, Drame, Lhoest, and Rush}]{wolf2020transformers}
Thomas Wolf, Lysandre Debut, Victor Sanh, Julien Chaumond, Clement Delangue,
  Anthony Moi, Pierric Cistac, Tim Rault, R\'{e}mi Louf, Morgan Funtowicz, Joe
  Davison, Sam Shleifer, Patrick von Platen, Clara Ma, Yacine Jernite, Julien
  Plu, Canwen Xu, Teven~Le Scao, Sylvain Gugger, Mariama Drame, Quentin Lhoest,
  and Alexander~M. Rush. 2020.
\newblock \href {https://aclanthology.org/2020.emnlp-demos.6/} {{Transformers:
  State-of-the-Art Natural Language Processing}}.
\newblock In \emph{Conference on Empirical Methods in Natural Language
  Processing \CNFX{EMNLP} - System Demonstrations}.

\bibitem[{Xie et~al.(2022)Xie, Wu, Shi, Zhong, Scholak, Yasunaga, Wu, Zhong,
  Yin, Wang et~al.}]{xie2022unifiedskg}
Tianbao Xie, Chen~Henry Wu, Peng Shi, Ruiqi Zhong, Torsten Scholak, Michihiro
  Yasunaga, Chien-Sheng Wu, Ming Zhong, Pengcheng Yin, Sida~I Wang, et~al.
  2022.
\newblock \href {https://arxiv.org/abs/2201.05966} {{UnifiedSKG: Unifying and
  Multi-Tasking Structured Knowledge Grounding with Text-to-Text Language
  Models}}.
\newblock \emph{arXiv preprint arXiv:2201.05966}.

\bibitem[{Xu et~al.(2022)Xu, Chen, Du, Shao, Wang, Li, and
  Yang}]{xu2022zeroprompt}
Hanwei Xu, Yujun Chen, Yulun Du, Nan Shao, Yanggang Wang, Haiyu Li, and Zhilin
  Yang. 2022.
\newblock \href {https://arxiv.org/abs/2201.06910} {{ZeroPrompt: Scaling
  Prompt-Based Pretraining to 1,000 Tasks Improves Zero-Shot Generalization}}.
\newblock \emph{arXiv preprint arXiv:2201.06910}.

\bibitem[{Xue et~al.(2021)Xue, Constant, Roberts, Kale, Al-Rfou, Siddhant,
  Barua, and Raffel}]{xue2021mt5}
Linting Xue, Noah Constant, Adam Roberts, Mihir Kale, Rami Al-Rfou, Aditya
  Siddhant, Aditya Barua, and Colin Raffel. 2021.
\newblock \href {https://aclanthology.org/2021.naacl-main.41/} {{mT5: A
  Massively Multilingual Pre-trained Text-to-Text Transformer}}.
\newblock In \emph{Conference of the North American Chapter of the Association
  for Computational Linguistics \CNFX{NAACL}}.

\bibitem[{Ye et~al.(2021)Ye, Lin, and Ren}]{ye2021crossfit}
Qinyuan Ye, Bill~Yuchen Lin, and Xiang Ren. 2021.
\newblock \href {https://aclanthology.org/2021.emnlp-main.572/} {{CrossFit: A
  Few-shot Learning Challenge for Cross-task Generalization in NLP}}.
\newblock In \emph{Conference on Empirical Methods in Natural Language
  Processing \CNFX{EMNLP}}.

\bibitem[{Ye and Ren(2021)}]{ye2021learning}
Qinyuan Ye and Xiang Ren. 2021.
\newblock \href {https://aclanthology.org/2021.acl-short.82/} {{Learning to
  Generate Task-Specific Adapters from Task Description}}.
\newblock In \emph{Annual Meeting of the Association for Computational
  Linguistics \CNFX{ACL}}.

\bibitem[{Zhong et~al.(2021)Zhong, Lee, Zhang, and Klein}]{zhong2021adapting}
Ruiqi Zhong, Kristy Lee, Zheng Zhang, and Dan Klein. 2021.
\newblock \href {https://aclanthology.org/2021.findings-emnlp.244/} {{Adapting
  Language Models for Zero-shot Learning by Meta-tuning on Dataset and Prompt
  Collections}}.
\newblock In \emph{Conference on Empirical Methods in Natural Language
  Processing \CNFX{EMNLP} - Findings}.

\end{thebibliography}
\bibliographystyle{acl_natbib}

\appendix
\clearpage

\begin{center}
{\Large \textbf{Supplemental Material}}
\end{center}

\section{Crowdsourcing Human Feedback}
\label{sec:crowdsourced:feedback}
We use Amazon Mechanical Turk (AMT) to crowdsource feedback on the quality of the collected instructions. 
We limit our crowdworkers to predominantly English-speaking countries (USA, UK, Canada, and Australia), and to those who have finished over 1K HITs with an approval rating of over $99\%$. 

\begin{figure*}[t!]
    \centering
    \fbox{\includegraphics[scale=0.26,trim=0.2cm 1.4cm 0.8cm 1.4cm,clip=true]{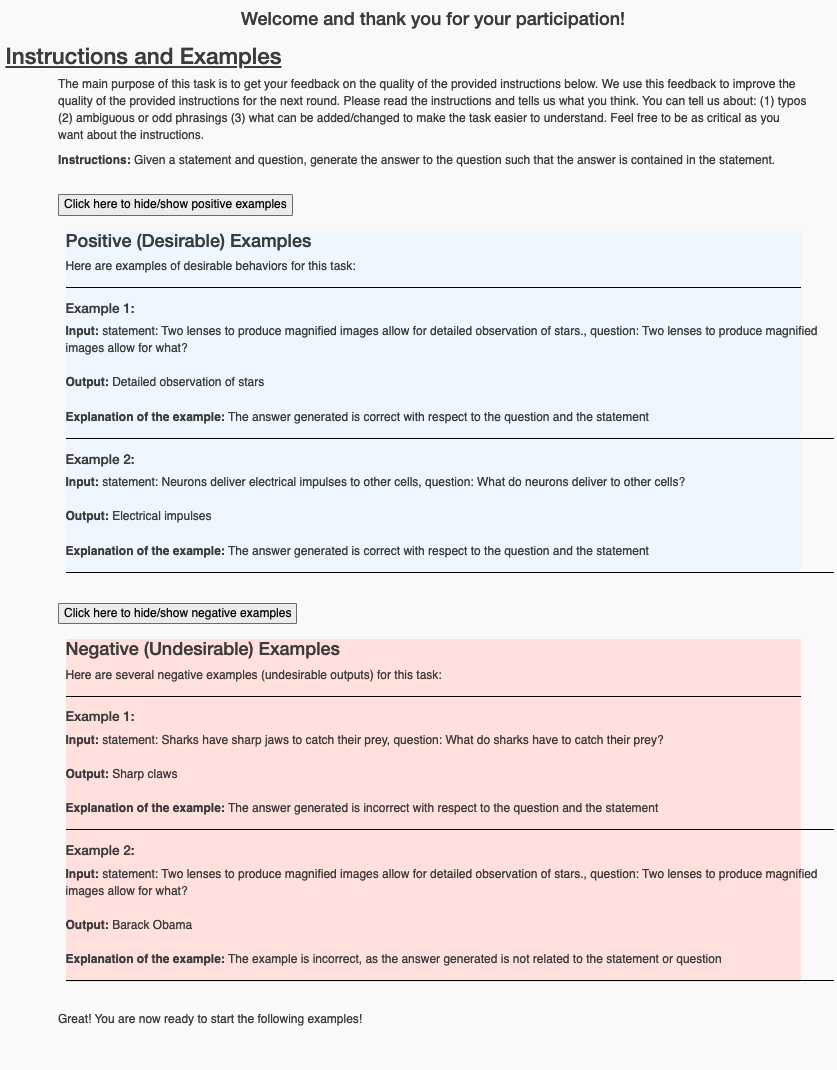}}
    \fbox{\includegraphics[scale=0.26,trim=0.2cm 4.71cm 1.2cm 0.4cm,clip=true]{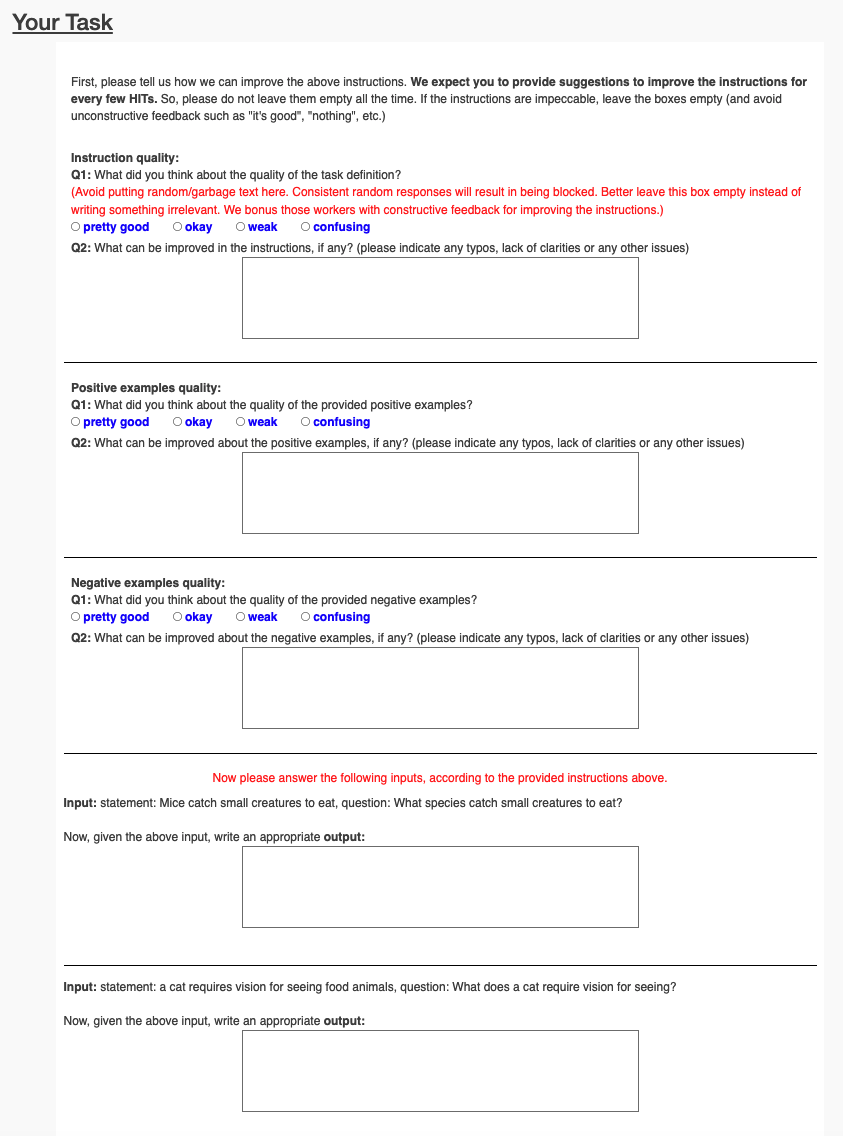}}
    \caption{The crowdsourcing template we use to receive feedback on our collected tasks.}
    \label{fig:crowdsourcing:feedback1}
\end{figure*}

Fig.~\ref{fig:crowdsourcing:feedback1} shows the crowdsourcing template used for collecting crowdworker feedback on our instructions. 
We show the instructions (the task definition, along with positive and negative examples) followed by forms for their feedback. 
We allow the crowdworkers to give us a qualitative measure of their perceived quality as well as text boxes for more concrete items (such as typos or phrasings that may benefit from more clear articulation). 
For each task, we solicit the feedback of 3 crowdworkers and then use this feedback to improve the task definitions or the examples for each task. 

\section{Crowdsourcing Human Judgements of Generation Quality}
\label{sec:crowdsourcing:judgements}
\begin{figure*}[t!]
    \centering
    \fbox{\includegraphics[scale=0.25,trim=0.8cm 3.4cm 2.3cm 1.9cm,clip=true]{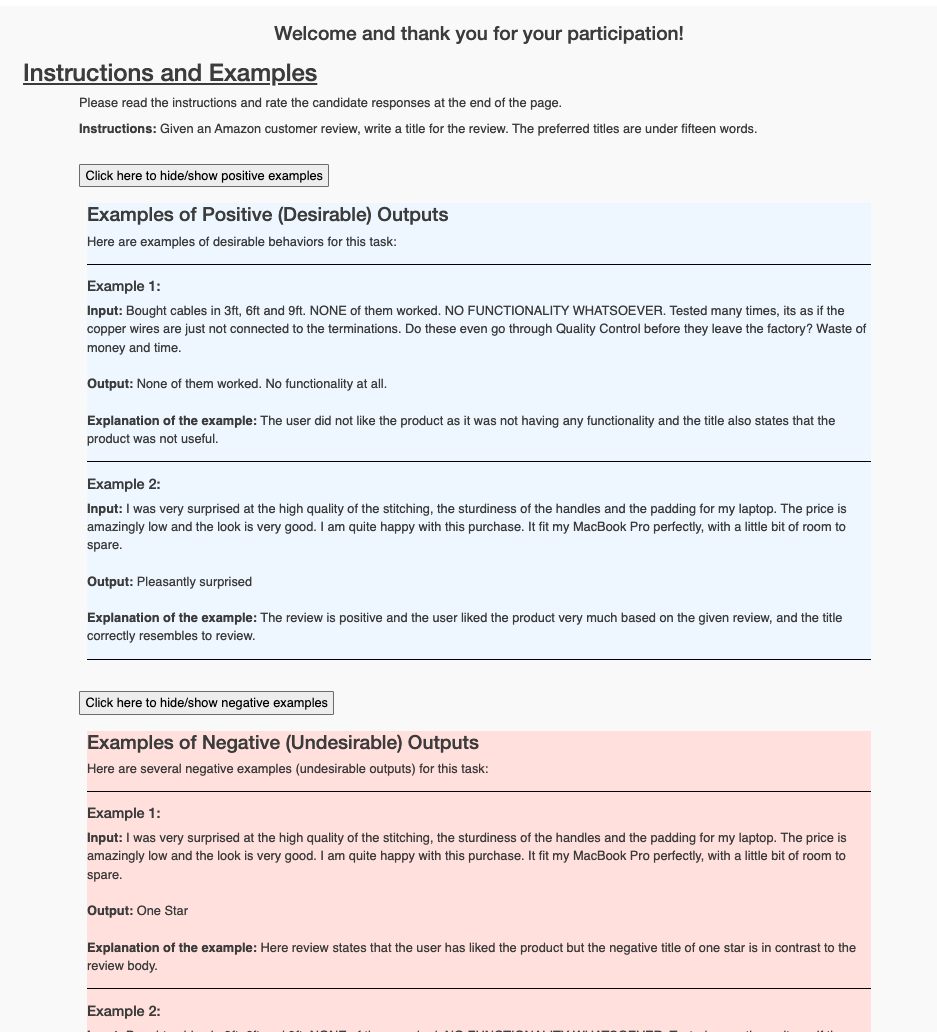}}
    \fbox{\includegraphics[scale=0.25,trim=0.6cm 4.71cm 2.5cm 0.4cm,clip=true]{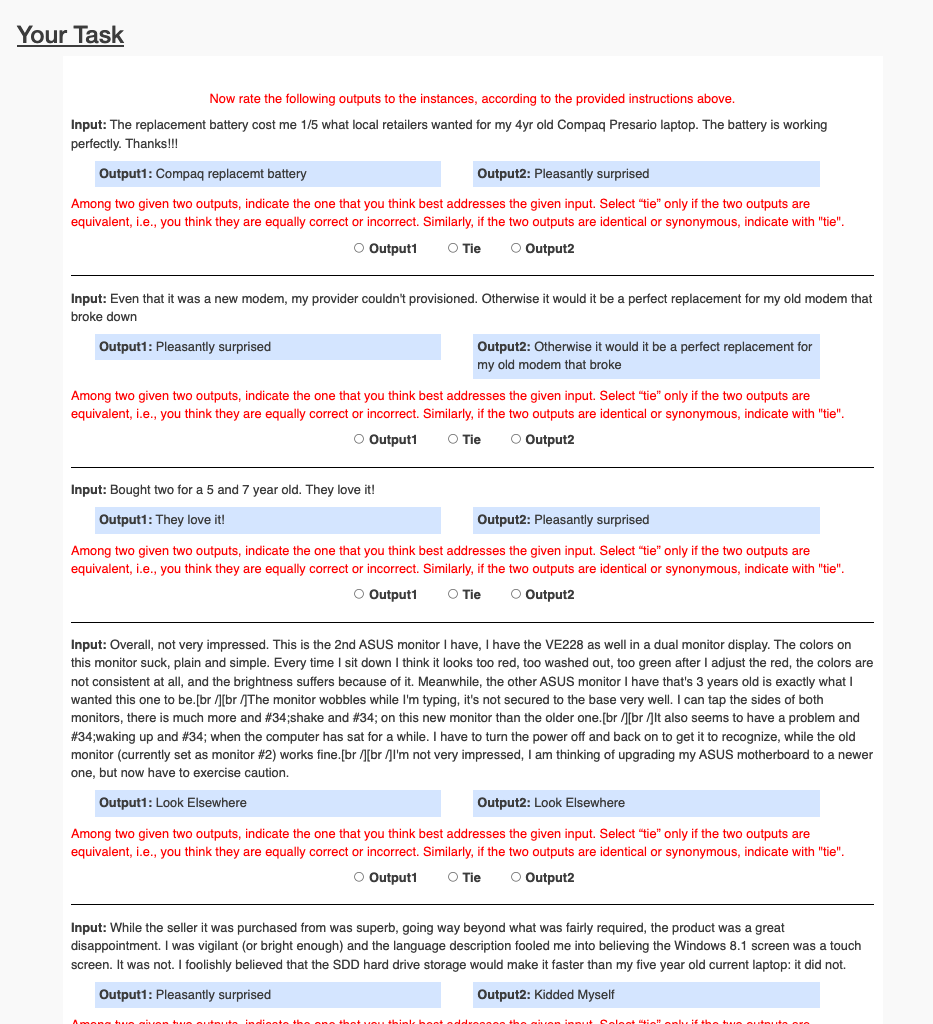}}
    \caption{Crowdsourcing interface used for human assessment of our baselines (\S\ref{subsec:human_eval}). 
    }
    \label{fig:crowdsourcing:template}
\end{figure*}

We perform a crowdsourcing experiment on Amazon Mechanical Turk (AMT) to assess the quality of the generated responses of models. 
Specifically, we ask crowdworkers to indicate if they prefer the predicted answer by the model or the ground truth outputs for each instances. 
The annotation interface is shown in Fig.~\ref{fig:crowdsourcing:template}. It is essentially the same template used for the quality assessment of the dataset (\S\ref{sec:crowdsourced:feedback}), except that here the crowdworkers are shown a \emph{pair} of responses for each instances---the reference text (from our benchmark) and the one generated by the model---turning the task into a \emph{comparative} evaluation. 

For each instance, we obtain annotations from an annotator as to whether they prefer either response over the other or they would rate them equally (``tie'').  The model receives a credit of 1.0 if the worker favors the model's prediction at least as well as the ground truth label (otherwise, the model would receive a credit of 0.0).   
The overall accuracy score for the model is computed by averaging instance-level scores.  
To reduce the costs, the human evaluation of our models is done on 60 randomly selected tasks (about half of our evaluation tasks), and on 10 random instances of each task. 

Since it is non-trivial to find non-English speaking crowdworkers~\cite{pavlick2014language}, this evaluation was restricted to English language tasks. 
Therefore, since our task is focused on English tasks, we required workers to be based in a country with a population predominantly of native English speakers (USA, Canada, UK, and Australia) and have completed at least 5000 HITs with $\geq$99\% assignment approval rate.

The resulting human-evaluation metric indicates how often were model predictions equal or preferred to our ground truth labels. 
In this evaluation, the theoretical upper bound is 100\% where the model is rated at least as well as the ground truth. 
The results of human evaluation are shown in the bottom row of Fig.~\ref{fig:human-eval-results}.

\section{Instruction Schema}
\label{sec:comparison:schema}

Our instruction schema is based on that of \naturalinsShort{} \cite{mishra2022cross}, but we simplify it to make data collection easier. 
Our \underline{\textsc{Definition}} field serves as the union of  \citet{mishra2022cross}'s \underline{\textsc{Definition}}, \underline{\textsc{Things to Avoid}}, and \underline{\textsc{Emphasis \& Caution}}. 
Additionally, we drop their \underline{\textsc{Title}} and \underline{\textsc{Prompt}} as their content is most often covered by  \underline{\textsc{Definition}}.

\section{Model Implementation Details}
\label{sec:implementation}

\paragraph{T5 experiments.} We use T5 for training our \modelName{}, estimating the performance of the supervised approach and conducting analysis. 

Our experiments that finetune the T5-11B model are conducted based on the Google's T5 library\footnote{\url{https://github.com/google-research/text-to-text-transfer-transformer}} and we use their T5.1.1.xxl checkpoint\footnote{\url{https://console.cloud.google.com/storage/browser/t5-data/pretrained_models/t5.1.1.xxl}} by default, which is pre-trained only on C4.\footnote{We also tried to finetune \modelName{} from \href{https://console.cloud.google.com/storage/browser/t5-data/pretrained_models/t5.1.1.lm100k.xxl?pli=1}{the T5-LM checkpoint} but the final performance is worse. Therefore, we decided to use the T5.1.1.xxl checkpoint.}
These experiments are run on Google V3-256 TPUs using a batch size of 1,048,576 tokens (1,024 examples), a constant learning rate of 1e-5 and a total of 1000 steps. Each training run takes 4 hours to complete.

Our analyses that use T5 models smaller than 11B parameters are conducted based on Huggingface's transformers library and model checkpoints\footnote{\url{https://huggingface.co/models?sort=downloads&search=google\%2Ft5}} \cite{wolf2020transformers} on GPU machines. When fine-tuning models, we train them for two epochs with a batch size of 16 and a constant learning rate of 1e-5. The maximum input length is set to 1024, and the maximum output length is set to 128. These experiments are conducted with 8 A100 GPUs with 48GB GPU memory per each. We use DeepSpeed\footnote{\url{https://github.com/microsoft/DeepSpeed}} for model parallelization, with bfloat16 precision enabled to save the GPU memory. Each training run takes 6 hours to complete.



\paragraph{GPT-3 and InstructGPT experiments.}
We use the OpenAI API\footnote{\url{https://beta.openai.com/docs/introduction/overview}} for conducting the GPT-3 experiments. We use their ``davinci'' engine for the GPT-3 language model experiments and their ``text-davinci-001'' engine for the InstructGPT experiments. When making the requests, we set the temperature as 0, top\_p as 1 and the maximum generation length as 128. Due to the high cost, we randomly sample 20 instances from each of our 119 test tasks to estimate the performance of GPT-3 and InstructGPT. All API requests were made on 
May 30, 2022.

\paragraph{Encoding instruction with input} 
\label{subsec:encodings} 
For every problem setup, we map a given instruction $I_t$ and an input instance $x$ into a textual format, obtaining $enc(I_t, x)$. 
Each instruction $I_t$ consists of multiple elements as described in our instruction schema (\S\ref{sec:dataset}). We map each element of the instruction to a textual format and prepend it to the input instance.  Fig.~\ref{fig:encoding} shows how we encode the full instruction. We study different combinations of these instruction elements in \S \ref{subsec:effects_of_encoding}. 
The encoded instance is then fed to an encoder-decoder model to predict $y$: $M:enc(I_t, x) \rightarrow y$. 

\setlength{\jot}{0.7pt} 
\begin{figure}[h]
\centering
\begin{boxedminipage}{0.9\columnwidth}
\vspace{-0.3cm}
\begin{equation*} 
    \scriptsize
    \begin{split}
        \hspace{0.5cm} & \mathtt{\small Definition: \;   \color{purple} \{\{definition\}\} }  \\  
        & \mathtt{\small Positive\; Example\;1-}   \\ 
        & \hspace{0.8cm} \mathtt{\small input: \color{purple} \{\{p\_ex1.input\}\}} \; \\ 
        & \hspace{0.8cm} \mathtt{\small output: \color{purple} \{\{p\_ex1.output\}\}} \; \\ 
        & \hspace{0.8cm} \mathtt{\small explanation: \color{purple} \{\{p\_ex1.exp\}\}} \; \\ 
        & \mathtt{\small Positive\; Example\;2-}   \\ 
        & \hspace{0.8cm}  \cdots  \\ 
        & \textnormal{}\mathtt{\small Negative\; Example \;1-}   \\ 
        & \hspace{0.8cm} \mathtt{\small input: \color{purple} \{\{n\_ex1.input\}\}} \; \\ 
        & \hspace{0.8cm} \mathtt{\small output: \color{purple} \{\{n\_ex1.output\}\}} \; \\ 
        & \hspace{0.8cm} \mathtt{\small explanation: \color{purple} \{\{n\_ex1.exp\}\}} \; \hspace{3.3cm} \\ 
        & \mathtt{\small Negative\; Example\;2-}   \\ 
        & \hspace{0.8cm}  \cdots  \\ 
        & \mathtt{\small Now \; complete \; the \; following \; example -} \; \\ 
        & \hspace{0.8cm}  \mathtt{\small input: \; \color{purple} \{\{x.input\}\} }   \\ 
        & \hspace{0.8cm}  \mathtt{\small output:} 
    \end{split}
\end{equation*}
\end{boxedminipage}
\caption{Encoding task instruction with input. 
}
\label{fig:encoding}
\end{figure}

\section{Evaluation Metrics}
\label{sec:metric-correlation}

\begin{figure*}[ht]
    \centering
    \includegraphics[width=\textwidth, trim=0cm 0cm 0cm 0cm]{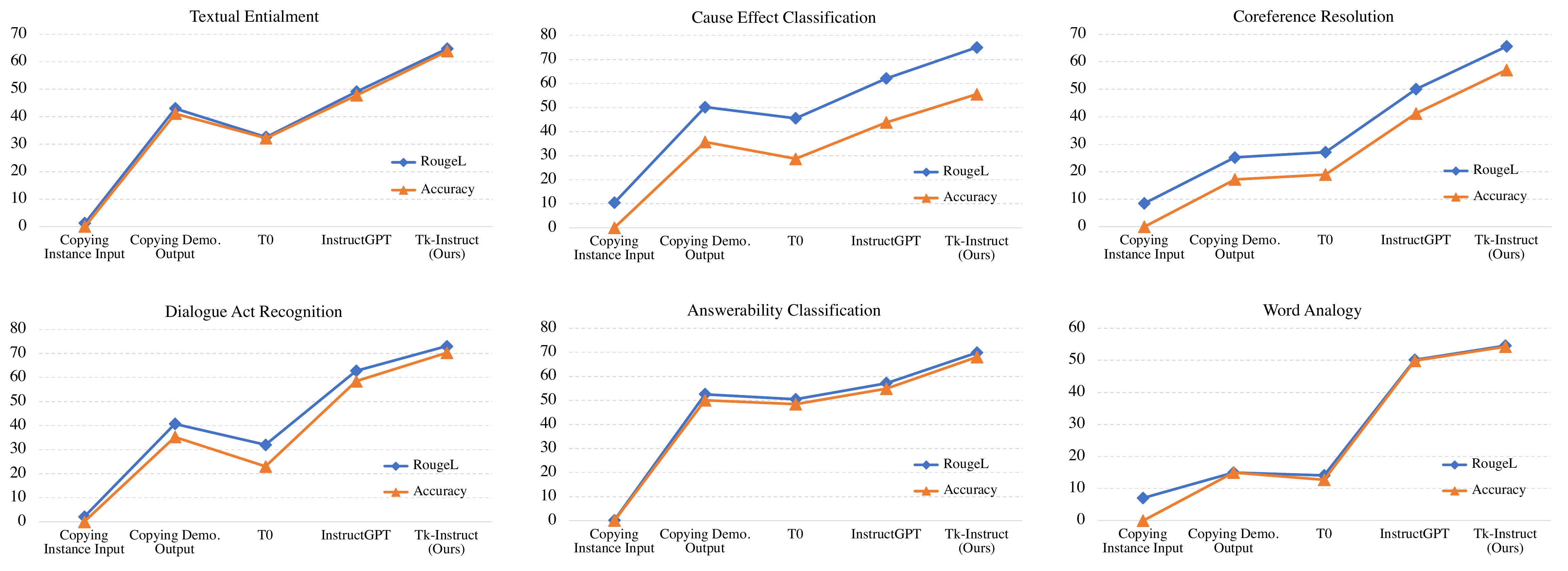}
    \caption{Rouge-L v.s. Accuracy for task types that are usually regarded as classification tasks. The trends of these two metrics are highly correlated with a Pearson coefficient of 0.970. 
    }
    \label{fig:rouge-em-correlation}
\end{figure*}

We adopt ROUGE-L as our automatic evaluation metric in this work. However, it remains a question for how much ROUGE-L can reflect model's performance on different tasks. Although we cannot test ROUGE-L's correlation with each task-specific metric of the tasks included in our data, we do investigate whether ROUGE-L can be used for classification tasks. Fig. \ref{fig:rouge-em-correlation} plots the ROUGE-L scores and accuracy of several models on different types of tasks. These task types are usually regarded as classification tasks and have very short ground truth output. We can see that for all these task types, the trend of ROUGE-L correlates well with the trend of accuracy. For some task types, we do see some gap between these two metrics. The reason is because there are some generation tasks categorized into these types. These results indicate that ROUGE-L is a good proxy for accuracy for classification tasks.

\section{Distribution of Tasks}
\label{sec:distribution}

As is described in \S\ref{sec:dataset}, \nameShort{} provides the annotation for categorizing tasks along three different dimensions: task type, language, and domain.  Fig.~\ref{fig:histograms} shows the distribution of tasks among these three dimensions. This meta-information can be used to study model's generalization ability in different senses. 
Despite the diversity of the data, we acknowledge the skew toward certain tasks and languages, which we leave to be addressed by future work.

\newcommand{\TempSubtable}[2][\empty]
{\hbox{
  \sbox0{#2}
  \captionsetup{position=top, justification=raggedleft, singlelinecheck=false, font=scriptsize}%
  \rotatebox[origin=bl]{90}{\begin{minipage}[b]{\dimexpr \ht0+\dp0}
    \subcaption{#1}
  \end{minipage}}\raisebox{\dp0}{\usebox0}%
}}
\begin{figure*}[t!]
\hspace{-2pt}
        \TempSubtable[Task Types\label{fig:Categories}]{%
       \includegraphics[width=0.98\textwidth]{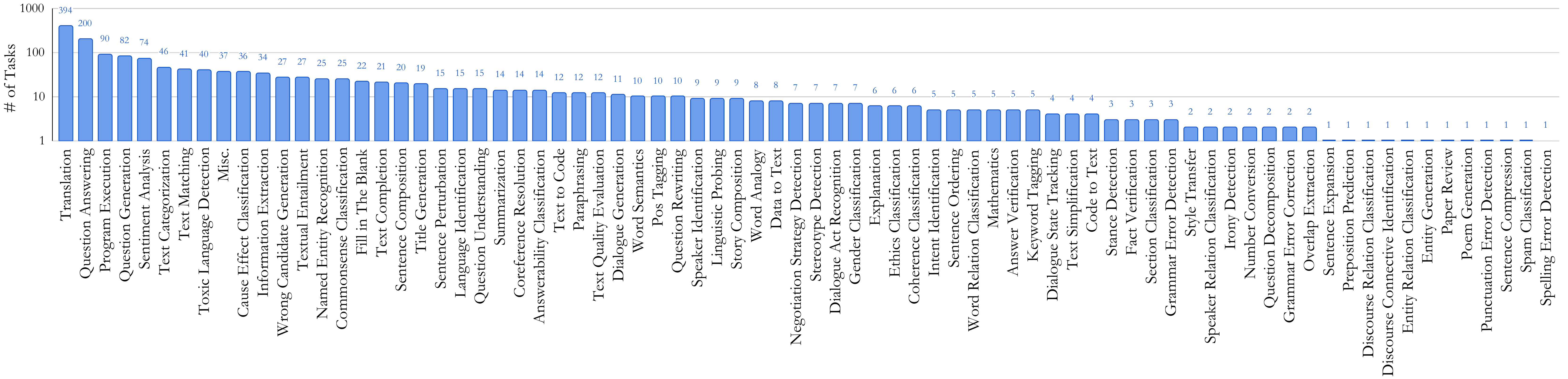}%
    }
    \TempSubtable[Languages\label{fig:Languages}]{%
       \includegraphics[width=0.98\textwidth]{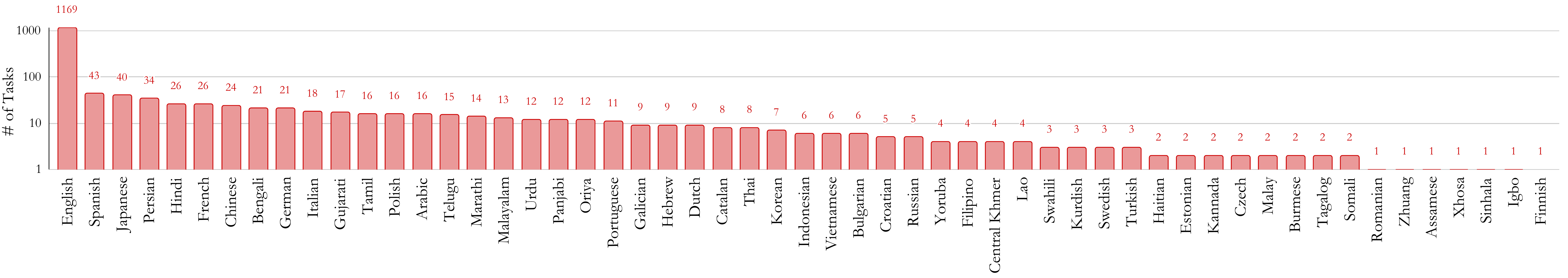}%
    }
        \TempSubtable[Domains\label{fig:Domains}]{%
       \includegraphics[width=0.98\textwidth]{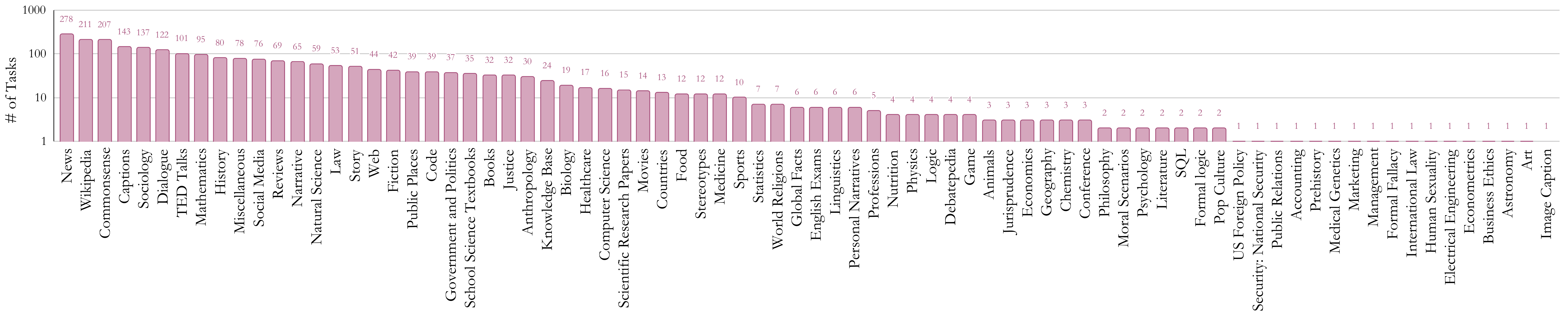}%
    }
    
    \caption{Distribution of \nameShort{} tasks in terms of their (a) task types (b) languages (c) domains.
    $y$-axes are in log scale. 
    }
    \label{fig:histograms}
\end{figure*} 

\section{Evaluation Tasks}
\label{sec:eval tasks}


Table~\ref{tab:test_categories} lists the 12 task categories used for our evaluation and all the tasks included in each category (introduced in \S\ref{subsec:setup}). To provide a better sense of what those tasks look like, we also select one representative task from each category and list them in Tables \ref{example:textual_entailment}--\ref{example:grammar_error_correction}. Due to the large number of tasks in our dataset, we cannot list all 1,616 tasks in this paper. We refer the reader to our dataset. 

\begin{table*}[ht]
\centering
\scriptsize
\resizebox{\textwidth}{!}{%
\begin{tabular}{p{0.1\textwidth}|p{0.1\textwidth}|p{0.4\textwidth}|p{0.4\textwidth}}
\toprule
Task Category & Metric & \multicolumn{2}{c}{List of Tasks} \\ \hline
Textual \newline Entailment & Exact Match & task937\_defeasible\_nli\_atomic\_textual\_entailment \; task202\_multinli\_textual\_entailment \; task936\_defeasible\_nli\_atomic\_textual\_entailment \; task641\_e\_snli\_textual\_entailment \; task1344\_rte\_textual\_entailment \; task1615\_sick\_textual\_entailment \; task1385\_anli\_textual\_entailment \; task935\_defeasible\_nli\_atomic\_textual\_entailment \; task199\_multinli\_textual\_entailment \; task1388\_cb\_textual\_entailment \;  task1554\_scitail\_textual\_entailment \; task640\_e\_snli\_textual\_entailment \; task534\_farstail\_textual\_entailment \; task201\_multinli\_textual\_entailment \; & task1386\_anli\_textual\_entailment \; task463\_pasinlu\_textual\_entailment \; task1387\_anli\_textual\_entailment \; task738\_perspectrum\_textual\_entailment \; task1529\_scitailv1.1\_textual\_entailment \; task190\_snli\_textual\_entailment \; task200\_multinli\_textual\_entailment \; task1612\_sick\_textual\_entailment \; task970\_sherliic\_textual\_entailment \; task890\_gwsd\_textual\_entailment \; task464\_pasinlu\_textual\_entailment \; task1516\_imppres\_textual\_entailment \; task642\_e\_snli\_textual\_entailment \\ \hline
Cause \newline Effect \newline Classification & Exact Match & task1178\_xcopa\_cause\_effect\_classification \; task391\_cod3s\_cause\_effect\_classification \; task939\_indicnlp\_cause\_effect\_classification \; task392\_cod3s\_cause\_effect\_classification \; task938\_indicnlp\_cause\_effect\_classification \; task1168\_xcopa\_cause\_effect\_classification \; task828\_copa\_cause\_effect\_classification \; task1628\_copa\_hr\_cause\_effect\_classification \; task943\_indicnlp\_cause\_effect\_classification \; task1182\_xcopa\_cause\_effect\_classification \; task1171\_xcopa\_cause\_effect\_classification \; task968\_xcopa\_cause\_effect\_classification \; task942\_indicnlp\_cause\_effect\_classification \; task1181\_xcopa\_cause\_effect\_classification \; task1172\_xcopa\_cause\_effect\_classification \; task1393\_copa\_cause\_effect\_classification \; task1174\_xcopa\_cause\_effect\_classification \;  task1627\_copa\_hr\_cause\_effect\_classification \; task1177\_xcopa\_cause\_effect\_classification \; &  task1184\_xcopa\_cause\_effect\_classification \; task1185\_xcopa\_cause\_effect\_classification \; task1176\_xcopa\_cause\_effect\_classification \; task614\_glucose\_cause\_effect\_classification \; task1629\_copa\_hr\_cause\_effect\_classification \; task1175\_xcopa\_cause\_effect\_classification \; task827\_copa\_cause\_effect\_classification \; task1173\_xcopa\_cause\_effect\_classification \; task1180\_xcopa\_cause\_effect\_classification \; task1170\_xcopa\_cause\_effect\_classification \; task1183\_xcopa\_cause\_effect\_classification \; task969\_xcopa\_cause\_effect\_classification \; task941\_indicnlp\_cause\_effect\_classification \; task1626\_copa\_hr\_cause\_effect\_classification \; task940\_indicnlp\_cause\_effect\_classification \; task393\_cod3s\_cause\_effect\_classification \; task1169\_xcopa\_cause\_effect\_classification \; task1179\_xcopa\_cause\_effect\_classification \\ \hline
Coreference Resolution & Exact Match & task1391\_winogrande\_coreference\_resolution \; task1664\_wino\_bias\_coreference\_resolution \; task304\_numeric\_fused\_head\_coreference\_resolution \; task892\_gap\_coreference\_resolution \; task891\_gap\_coreference\_resolution \; task330\_gap\_coreference\_resolution \; task401\_numeric\_fused\_head\_coreference\_resolution \; & task033\_winogrande\_coreference\_resolution \; task133\_winowhy\_coreference\_resolution \; task329\_gap\_coreference\_resolution \; task249\_enhanced\_wsc\_coreference\_resolution \; task648\_winograd\_wsc\_coreference\_resolution \; task1390\_wsc\_fiexed\_coreference\_resolution \; task893\_gap\_coreference\_resolution \\ \hline
Dialogue \newline Act \newline Recognition & Exact Match & task879\_schema\_guided\_dstc8\_dialogue\_act\_recognition \; task362\_spolin\_dialogue\_act\_recognition \;  task1533\_dailydialog\_dialogue\_act\_recognition \;  task1534\_dailydialog\_dialogue\_act\_recognition \;  & task880\_schema\_guided\_dstc8\_dialogue\_act\_recognition \; task1531\_dailydialog\_dialogue\_act\_recognition \; task1394\_meta\_woz\_dialogue\_act\_recognition \\ \hline
Answerability Classification & Exact Match & task020\_mctaco\_answerability\_classification \; task050\_multirc\_answerability\_classification \; task1439\_doqa\_answerability\_classification \; task233\_iirc\_answerability\_classification \; task226\_stack\_overflow\_answerability\_classification \; task396\_persianqa\_answerability\_classification \; task1640\_adverserial\_qa\_answerability\_classification \; & task232\_iirc\_answerability\_classification \; task1442\_doqa\_answerability\_classification \; task242\_tweetqa\_answerability\_classification \; task1624\_disfl\_qa\_answerability\_classification \; task520\_aquamuse\_answerability\_classification \; task290\_tellmewhy\_answerability\_classification \; task349\_squad2.0\_answerability\_classification \\ \hline
Word Analogy & Exact Match & task1155\_bard\_word\_analogy \; task1152\_bard\_word\_analogy \; task1158\_bard\_word\_analogy \; task1156\_bard\_word\_analogy \; task1157\_bard\_word\_analogy \; & task1159\_bard\_word\_analogy \; task1153\_bard\_word\_analogy \; task1154\_bard\_word\_analogy \\ \hline
Overlap \newline Extraction & ROUGE-L & task039\_qasc\_overlap\_extraction \; & task281\_points\_of\_correspondence\_overlap\_extraction \\ \hline
Keyword \newline Tagging & ROUGE-L & task613\_liar\_keyword\_tagging \; task645\_wiki\_auto\_all\_data\_keyword\_tagging \; task620\_ohsumed\_keyword\_tagging \; & 
task036\_qasc\_keyword\_tagging \; task623\_ohsumed\_keyword\_tagging \\ \hline
Question Rewriting & ROUGE-L & task670\_ambigqa\_question\_rewriting \; task121\_zest\_question\_rewriting \; task1195\_disfl\_qa\_question\_rewriting \; task442\_com\_qa\_question\_rewriting \; 
task1345\_qqp\_question\_rewriting \;  task035\_winogrande\_question\_rewriting \; & task671\_ambigqa\_question\_rewriting \;   task1562\_zest\_question\_rewriting \; task1622\_disfl\_qa\_question\_rewriting \; task034\_winogrande\_question\_rewriting \; task402\_grailqa\_question\_rewriting \\ \hline
Title \newline Generation & ROUGE-L & task1356\_xlsum\_title\_generation \; task1540\_peer\_read\_title\_generation \; task1659\_billsum\_title\_generation \; task569\_recipe\_nlg\_title\_generation \; task1342\_amazon\_us\_reviews\_title\_generation \; task220\_rocstories\_title\_generation \; task1561\_clickbait\_news\_bg\_title\_generation \; task418\_persent\_title\_generation \; task1358\_xlsum\_title\_generation \; task769\_qed\_title\_generation \; & 
task219\_rocstories\_title\_generation \;  task602\_wikitext\_title\_generation \; task1586\_scifact\_title\_generation \; task743\_eurlex\_title\_generation \; task500\_scruples\_title\_generation \; task619\_ohsumed\_title\_generation \; task510\_reddit\_tifu\_dataset\_title\_generation \; task288\_gigaword\_title\_generation \; task1161\_coda\_19\_title\_generation \\ \hline
Data to Text & ROUGE-L & task957\_e2e\_data\_to\_text \; task1631\_open\_pi\_data\_to\_text \; task1598\_nyc\_data\_to\_text \; task1728\_web\_nlg\_data\_to\_text \; task102\_commongen\_data\_to\_text \; &  task677\_ollie\_data\_to\_text \; task1407\_dart\_data\_to\_text \; task1409\_dart\_data\_to\_text \; task760\_msr\_sqa\_data\_to\_text \\ \hline
Grammar Error Correction & ROUGE-L &  task1557\_jfleg\_grammar\_error\_correction & \\
\bottomrule
\end{tabular}%
}
\caption{12 Evaluation categories (\S\ref{subsec:setup}), their evaluation metrics (Exact Matching or ROUGE-L, \S\ref{subsec:metrics}), and all the tasks in each category.}
\label{tab:test_categories}
\end{table*}

\begin{table*}[ht]
\centering
\small
\resizebox{0.9\textwidth}{!}{%
\begin{tabular}{p{0.1\textwidth}|p{0.9\textwidth}}
\toprule
Task Type & Textual Entailment \\ \hline
Task ID & task1344\_rte\_textual\_entailment \\ \hline
Definition &
  In this task, you're given two sentences. Indicate if the first sentence clearly entails the second sentence (i.e., one can conclude the 2nd sentence by  reading the 1st one). Indicate your answer with ``1'' if the first sentence entails the second sentence, otherwise answer with ``0''. 
   \\ \hline  
Positive Example & 
  \textbf{Input}: Sentence 1: No Weapons of Mass Destruction Found in Iraq Yet. Sentence 2:Weapons of Mass Destruction Found in Iraq. \newline 
  \textbf{Output}: 0 \newline 
  \textbf{Explanation}: In our  first statement we clearly say that Iraq does not have any weapon of mass destruction but the second sentence says that weapon of mass destruction is found in Iraq which is a contradiction. Hence output will be 0 for non entailment.
   \\ \hline 
Negative Example & 
  \textbf{Input}: Sentence 1: Valero Energy Corp., on Monday, said it found  "extensive" additional damage at its 250,000-barrel-per-day Port  Arthur refinery. Sentence 2: Valero Energy Corp. produces 250,000 barrels per day. \newline 
  \textbf{Output}: 0 \newline 
  \textbf{Explanation}: The first statement mentions that there was   damage found in the 250,000 barrel-per-day Port Aurthur refinery. Which means that they produce 250,000 barrels a day. Hence the output should have been 1   for entailment.
   \\ \hline
Instance & 
  \textbf{Input}: Sentence 1: Like the United States, U.N. officials are also dismayed that Aristide killed a conference called by Prime Minister Robert Malval in Port-au-Prince in hopes of bringing all the feuding parties together. Sentence 2: Aristide had Prime Minister Robert Malval  murdered in Port-au-Prince. \newline \textbf{Valid Output}: [``0'']
  \\
 \bottomrule
\end{tabular}%
}
\caption{An example task in the Textual Entailment category of our dataset, adopted from RTE~\cite{dagan2005pascal,bentivogli2008the}.
}
\label{example:textual_entailment}
\end{table*}

\begin{table*}[htbp]
\centering
\small
\resizebox{0.9\textwidth}{!}{%
\begin{tabular}{p{0.1\textwidth}|p{0.9\textwidth}}
\toprule
Task Type & Cause Effect Classification \\ \hline
Task ID & task828\_copa\_cause\_effect\_classification \\ \hline
Definition &
  In this task your given two statements. You must judge whether the second sentence is the cause or effect of the first one. Label the instances as ``cause'' or ``effect'' based on your judgment. The sentences are separated by a newline character. 
   \\ \hline  
Positive Example & 
  \textbf{Input}: The women met for coffee. They wanted to catch up with each other. \newline 
  \textbf{Output}: cause \newline 
  \textbf{Explanation}: The women met for coffee because they wanted to catch up with each other.
   \\ \hline 
Negative Example & 
  \textbf{Input}: My body cast a shadow over the grass. The sun was rising. \newline
  \textbf{Output}: effect \newline 
  \textbf{Explanation}: The rising of the sun isn't an effect of casting a shadow over the grass.
   \\ \hline
Instance & 
  \textbf{Input}: The woman tolerated her friend's difficult behavior. The woman knew her friend was going through a hard time. \newline 
  \textbf{Valid Output}: [``cause'']
  \\
 \bottomrule
\end{tabular}%
}
\caption{An example task in the Cause Effect Classification category of our dataset, adopted from COPA~\cite{roemmele2011choice}.}
\label{example:cause_effect_classification}
\end{table*}

\begin{table*}[htbp]
\centering
\small
\resizebox{0.9\textwidth}{!}{%
\begin{tabular}{p{0.1\textwidth}|p{0.9\textwidth}}
\toprule
Task Type & Coreference Resolution \\ \hline
Task ID & task1391\_winogrande\_coreference\_resolution \\ \hline
Definition &
  In this task, you are given a question containing a blank (\_) and two options. You should pick the best option to answer the question. Please answer with ``A'' or ``B''. 
   \\ \hline  
Positive Example & 
  \textbf{Input}: Katrina gave Christine a stuffed animal for their birthday, but \_ already had this one. (A) Katrina (B) Christine \newline 
  \textbf{Output}: B \newline 
  \textbf{Explanation}: Since the blank is someone who received the gift and already had a stuffed animal, the answer must be ``Christine''.
   \\ \hline 
Negative Example & 
  \textbf{Input}: Kevin had to use less sunscreen when at the beach tanning than Justin because \_ had less sensitive skin. (A) Kevin (B) Justin \newline
  \textbf{Output}: (A) \newline 
  \textbf{Explanation}: Here, an additonal parentheses has been added to the answer. Note that, a valid answer must be ``A'' or ``B''.
   \\ \hline
Instance & 
  \textbf{Input}: Benjamin hated being in the sand and just watched Nelson make castle since \_ hated to be messy. (A) Benjamin (B) Nelson \newline 
  \textbf{Valid Output}: [``A'']
  \\
 \bottomrule
\end{tabular}%
}
\caption{An example task in the Cause Effect Classification category of our dataset, adopted from WinoGrande~\cite{sakaguchi2020winogrande}.}
\label{example:coreference}
\end{table*}

\begin{table*}[htbp]
\centering
\small
\resizebox{0.9\textwidth}{!}{%
\begin{tabular}{p{0.1\textwidth}|p{0.9\textwidth}}
\toprule
Task Type & Dialogue Act Recognition \\ \hline
Task ID & task1394\_meta\_woz\_dialogue\_act\_recognition \\ \hline
Definition &
  In this task, you are given four sentences: a bot task sentence, a bot role sentence, a user task sentence and a user role sentence. Your job is to classify given sentences into one of the 47 different domains. The domains are: ``UPDATE\_CALENDAR'', ``PRESENT\_IDEAS'', ``MOVIE\_LISTINGS'', ``AUTO\_SORT'', ``GAME\_RULES'', ``CONTACT\_MANAGER'', ``BANK\_BOT'', ``MUSIC\_SUGGESTER'', ``CHECK\_STATUS'', ``PET\_ADVICE'', ``HOW\_TO\_BASIC'', ``NAME\_SUGGESTER'', ``QUOTE\_OF\_THE\_DAY\_BOT'', ``GUINESS\_CHECK'', ``INSURANCE'', ``RESTAURANT\_PICKER'', ``MAKE\_RESTAURANT\_RESERVATIONS'', ``WEDDING\_PLANNER'', ``SKI\_BOT'', ``HOME\_BOT'', ``PLAY\_TIMES'', ``BUS\_SCHEDULE\_BOT'', ``WHAT\_IS\_IT'', ``PHONE\_PLAN\_BOT'', ``DECIDER\_BOT'', ``PHONE\_SETTINGS'', ``TIME\_ZONE'', ``LIBRARY\_REQUEST'', ``UPDATE\_CONTACT'', ``CATALOGUE\_BOT'', ``PROMPT\_GENERATOR'', ``SCAM\_LOOKUP'', ``SPORTS\_INFO'', ``POLICY\_BOT'', ``CITY\_INFO'', ``APARTMENT\_FINDER'', ``EVENT\_RESERVE'', ``SHOPPING'', ``EDIT\_PLAYLIST'', ``LOOK\_UP\_INFO'', ``ORDER\_PIZZA'', ``WEATHER\_CHECK'', ``APPOINTMENT\_REMINDER'', ``GEOGRAPHY'', ``STORE\_DETAILS'', ``AGREEMENT\_BOT'', ``ALARM\_SET''.
   \\ \hline  
Positive Example & 
  \textbf{Input}: Bot's task: Inform the user that the topping they are asking for is unavailable. Bot's role: You are a bot designed to help customers order pizza. User's task: Ask if a certain pizza topping is available. User's role: You are interacting with a pizza restaurant bot. \newline 
  \textbf{Output}: ORDER\_PIZZA \newline 
  \textbf{Explanation}: According to the descriptions of the four sentences, we know that the type of task is ORDER\_PIZZA.
   \\ \hline 
Negative Example & 
  \textbf{Input}: Bot's task: Help the user with their pizza order. Bot's role: You are a bot designed to help customers order pizza. User's task: Ask the bot for three different pizzas. User's role: You are interacting with a pizza restaurant bot. \newline
  \textbf{Output}: UPDATE\_CALENDAR \newline 
  \textbf{Explanation}: According to the descriptions of the tasks and roles, we know that the type of task is ORDER\_PIZZA, but the output is UPDATE\_CALENDAR, so it is incorrect.
   \\ \hline
Instance & 
  \textbf{Input}: Bot's task: Tell the user when the movie is playing on Friday night. Bot's role: You are a a bot designed to provide movie listings. User's task: Ask the bot for the movie times for a movie on Friday night. User's role: You are interacting with a bot designed to provide movie listings. \newline 
  \textbf{Valid Output}: [``MOVIE\_LISTINGS'']
  \\
 \bottomrule
\end{tabular}%
}
\caption{An example task in the Dialogue Act Recognition category of our dataset, adopted from MetaLWOz~\cite{shalyminov2020fast}.}
\label{example:dialog_act}
\end{table*}

\begin{table*}[htbp]
\centering
\small
\resizebox{0.9\textwidth}{!}{%
\begin{tabular}{p{0.1\textwidth}|p{0.9\textwidth}}
\toprule
Task Type & Answerability Classification \\ \hline
Task ID & task1640\_adverserial\_qa\_answerability\_classification \\ \hline
Definition &
  Given a paragraph from a wikipedia article about some topic, and a question related to the topic, determine whether the question is answerable from the paragraph. If the question is answerable, answer ``True'', otherwise, answer ``False''.
   \\ \hline  
Positive Example & 
  \textbf{Input}: Another approach to brain function is to examine the consequences of damage to specific brain areas. Even though it is protected by the skull and meninges, surrounded by cerebrospinal fluid, and isolated from the bloodstream by the blood 2013 brain barrier, the delicate nature of the brain makes it vulnerable to numerous diseases and several types of damage. In humans, the effects of strokes and other types of brain damage have been a key source of information about brain function. Because there is no ability to experimentally control the nature of the damage, however, this information is often difficult to interpret. In animal studies, most commonly involving rats, it is possible to use electrodes or locally injected chemicals to produce precise patterns of damage and then examine the consequences for behavior. Question: What is surrounded by cerebrospinal fluid? \newline 
  \textbf{Output}: True \newline 
  \textbf{Explanation}: The paragraph comes from the wikipedia page on the brain. The answer to the question is the brain which can be found in the paragraph.
   \\ \hline 
Negative Example & 
  \textbf{Input}: NASCAR (headquartered in Daytona Beach) begins all three of its major auto racing series in Florida at Daytona International Speedway in February, featuring the Daytona 500, and ends all three Series in November at Homestead-Miami Speedway. Daytona also has the Coke Zero 400 NASCAR race weekend around Independence Day in July. The 24 Hours of Daytona is one of the world's most prestigious endurance auto races. The Grand Prix of St. Petersburg and Grand Prix of Miami have held IndyCar races as well. Question: What is the starting time of NASCAR's big events? \newline
  \textbf{Output}: False \newline 
  \textbf{Explanation}: This paragraph comes from the wikipedia article on Florida. The answer to the given question is February which can be found in the paragraph, however the output is given as False.
   \\ \hline
Instance & 
  \textbf{Input}: Another approach to brain function is to examine the consequences of damage to specific brain areas. Even though it is protected by the skull and meninges, surrounded by cerebrospinal fluid, and isolated from the bloodstream by the blood 2013 brain barrier, the delicate nature of the brain makes it vulnerable to numerous diseases and several types of damage. In humans, the effects of strokes and other types of brain damage have been a key source of information about brain function. Because there is no ability to experimentally control the nature of the damage, however, this information is often difficult to interpret. In animal studies, most commonly involving rats, it is possible to use electrodes or locally injected chemicals to produce precise patterns of damage and then examine the consequences for behavior. Question: What are the benefts of the blood brain barrier? \newline 
  \textbf{Valid Output}: [``True'']
  \\
 \bottomrule
\end{tabular}%
}
\caption{An example task in the Answerability Classification category of our dataset, adopted from AdversarialQA~\cite{bartolo2020beat}.}
\label{example:answerability}
\end{table*}

\begin{table*}[htbp]
\centering
\small
\resizebox{0.9\textwidth}{!}{%
\begin{tabular}{p{0.1\textwidth}|p{0.9\textwidth}}
\toprule
Task Type & Word Analogy \\ \hline
Task ID & task1156\_bard\_word\_analogy \\ \hline
Definition &
  Two analogies that relate actions to the tools used to perform the action is given in the form ``A : B. C : ?''. ``A : B'' relates action A to tool B. Your task is to replace the question mark (?) with the appropriate tool for the given action C, following the ``A : B'' relation.
   \\ \hline  
Positive Example & 
  \textbf{Input}: eat : fork. cook : ? \newline 
  \textbf{Output}: pan \newline 
  \textbf{Explanation}: The given analogy relates actions to the tools used to perform them. A fork can be used to eat. To cook, a pan can be used.
   \\ \hline 
Negative Example & 
  \textbf{Input}: dig : shovel. wash : ? \newline
  \textbf{Output}: sink \newline 
  \textbf{Explanation}: The given analogy relates actions to the tools used to perform them. A knife can be used to cut. To wash, a sink CANNOT be used.
   \\ \hline
Instance & 
  \textbf{Input}: cut : knife. wash : ? \newline 
  \textbf{Valid Output}: [``soap'', ``washcloth'', ``detergent'', ``rag'']
  \\
 \bottomrule
\end{tabular}%
}
\caption{An example task in the Word Analogy category of our dataset, adopted from BARD~\cite{fulda2017harvesting}.}
\label{example:word_analogy}
\end{table*}

\begin{table*}[htbp]
\centering
\small
\resizebox{0.9\textwidth}{!}{%
\begin{tabular}{p{0.1\textwidth}|p{0.9\textwidth}}
\toprule
Task Type & Overlap Extraction \\ \hline
Task ID & task281\_points\_of\_correspondence\_overlap\_extraction \\ \hline
Definition &
  You will be given three sentences. Read them, then identify a noun phrase (person, place, or thing) or event that is shared between all three sentences. As the output, write the span of the text corresponding to that phrase in each sentence. Keep the order of the sentences, that is, your answer should look like: 1: *a phrase from sentence 1* 2: *a phrase from sentence 2* 3: *a phrase from sentence 3*
   \\ \hline  
Positive Example & 
  \textbf{Input}: 1: Four employees of the store have been arrested , but its manager -- herself a woman -- was still at large Saturday , said Goa police superintendent Kartik Kashyap . 2: If convicted , they could spend up to three years in jail , Kashyap said . 3: The four store workers arrested could spend 3 years each in prison if convicted . \newline 
  \textbf{Output}: 1: Four employees of the store 2: they 3: The four store workers \newline 
  \textbf{Explanation}: All three mentioned parts refer to the same entity, the four employees. ``of the store'' in first sentence must be included, since it is part of the same noun phrase describing the employees.
   \\ \hline 
Negative Example & 
  \textbf{Input}: 1: But an Arizona official told CNN Bates never trained with the agency . 2:  He didn't come to Arizona ,  the official from the Maricopa County Sheriff 's Office said ,  and he certainly didn't train with us .  3: Maricopa County Sheriff 's Office in Arizona says Robert Bates never trained with them . \newline
  \textbf{Output}: 1: the agency 3: Maricopa County Sheriff 's Office in Arizona \newline 
  \textbf{Explanation}: The two noun phrases given in this example are correct, but there's no noun phrase from sentence 2. You should include all three sentences in your response.
   \\ \hline
Instance & 
  \textbf{Input}: 1: The President is headed to Panama for a regional summit , and Julie Pace of The Associated Press reports one of the big questions is whether he 'll make history and have a face-to-face meeting with Cuban leader Raul Castro . 2:  And so what the White House is going to be weighing is whether this meeting would be a way to generate more progress or whether it would be a premature reward for the Castros .  3: White House weighing whether Obama should meet with Raul Castro .\newline 
  \textbf{Valid Output}: [``1: Cuban leader Raul Castro 2: the Castros 3: Raul Castro'', ``1: face-to-face meeting 2: this meeting 3: meet'']
  \\
 \bottomrule
\end{tabular}%
}
\caption{An example task in the Overlap Extraction category of our dataset, adopted from PointsOfCorrespondence~\cite{lebanoff2020understanding}.}
\label{example:overlap}
\end{table*}

\begin{table*}[htbp]
\centering
\small
\resizebox{0.9\textwidth}{!}{%
\begin{tabular}{p{0.1\textwidth}|p{0.9\textwidth}}
\toprule
Task Type & Keyword Tagging \\ \hline
Task ID & task620\_ohsumed\_keyword\_tagging \\ \hline
Definition &
  Given an abstract, generate a keyword (a noun phrase) that best describes the focus or contribution of the paper. Such keywords can be directly from the given abstract or outside it.
   \\ \hline  
Positive Example & 
  \textbf{Input}: Abstract: Our results suggest that ethylene oxide retention after sterilization is increased in cuprammonium cellulose plate dialyzers containing potting compound. In contrast, cuprammonium cellulose plate dialyzers without potting compound were characterized by a rapid disappearance of retained ethylene oxide after sterilization. Whether these findings explain the low incidence of SARD with cuprammonium cellulose plate dialyzers that do not contain potting material is a matter for continued study and experimentation. \newline 
  \textbf{Output}: Sterilization \newline 
  \textbf{Explanation}: This term is directly present in the abstract and it is one of the main topic in it. So can be chosen as the medical subject heading.
   \\ \hline 
Negative Example & 
  \textbf{Input}: Abstract: Our results suggest that ethylene oxide retention after sterilization is increased in cuprammonium cellulose plate dialyzers containing potting compound. In contrast, cuprammonium cellulose plate dialyzers without potting compound were characterized by a rapid disappearance of retained ethylene oxide after sterilization. Whether these findings explain the low incidence of SARD with cuprammonium cellulose plate dialyzers that do not contain potting material is a matter for continued study and experimentation. \newline
  \textbf{Output}: Plasma Volume
  \newline 
  \textbf{Explanation}: This term is not directly present in the abstract and it is no way related to the abstract. So can not be chosen as the medical subject heading. ``Cellulose'' can be become a mesh term
   \\ \hline
Instance & 
  \textbf{Input}: Abstract: There is controversy regarding the appropriate utilization of health care resources in the management of tricyclic antidepressant overdosage. Antidepressant overdose patients presenting to the emergency department (ED) are routinely admitted to intensive care units, but only a small proportion develop cardiac arrhythmias or other complications requiring such an environment. The authors reviewed the findings in 165 patients presenting to an ED with antidepressant overdose. They found that major manifestations of toxicity on ED evaluation (altered mental status, seizures, arrhythmias, an
d conduction defects) were commonly associated with a complicated hospital course. Patients with the isolated findings of sinus tachycardia or QTc prolongation had no complications. No patient experienced a
 serious toxic event without major evidence of toxicity on ED evaluation and continued evidence of toxicity during the hospital course. These data support the concept that proper ED evaluation can identify
a large body of patients with trivial ingestions who may not require hospital observation.\newline 
  \textbf{Valid Output}: [``Antidepressive Agents, Tricyclic'',
        ``Arrhythmia'',
        ``California'',
        ``Electrocardiography'',
        ``Emergencies'',
        ``Emergency Service, Hospital'',
        ``Female'',
        ``Human'',
        ``Length of Stay'',
        ``Male'',
        ``Prognosis'',
        ``Retrospective Studies'',
        ``Tachycardia, Sinus'']
  \\
 \bottomrule
\end{tabular}%
}
\caption{An example task in the Keyword Tagging category of our dataset, adopted from OHSUMED~\cite{hersh1994ohsumed}.}
\label{example:keyword_tagging}
\end{table*}

\begin{table*}[htbp]
\centering
\small
\resizebox{0.9\textwidth}{!}{%
\begin{tabular}{p{0.1\textwidth}|p{0.9\textwidth}}
\toprule
Task Type & Question Rewriting \\ \hline
Task ID & task1622\_disfl\_qa\_question\_rewriting \\ \hline
Definition &
  Convert a disfluent question to a proper question. A disfluent question is a question that has some interruptions in it while framing. A proper question is the correct form of the question without any disfluency.
   \\ \hline  
Positive Example & 
  \textbf{Input}: Why was uh where was the Rhine regulated with an upper canal? \newline 
  \textbf{Output}: Where was the Rhine regulated with an upper canal? \newline 
  \textbf{Explanation}: The above disfluent question is correctly converted to a proper question. The 'uh' and the 'why' before the correction should be removed to make it fluent.
   \\ \hline 
Negative Example & 
  \textbf{Input}: When did Maududi exert the least impact no where did he exert? \newline
  \textbf{Output}: When did Maududi exert the least impact?
  \newline 
  \textbf{Explanation}: The above disfluent question is not correctly converted to a proper question since the original meaning in the disfluent question is where did Maudidi exert the least impact but the output asks when did Maudidi exert the least impact.
   \\ \hline
Instance & 
  \textbf{Input}: What kind of, no hold up, what describes the proportionality of acceleration to force and mass?\newline 
  \textbf{Valid Output}: [``What describes the proportionality of acceleration to force and mass?'']
  \\
 \bottomrule
\end{tabular}%
}
\caption{An example task in the Question Rewriting category of our dataset, adopted from Disfl-QA~\cite{gupta2021disfl}.}
\label{example:disflqa}
\end{table*}

\begin{table*}[htbp]
\centering
\small
\resizebox{0.9\textwidth}{!}{%
\begin{tabular}{p{0.1\textwidth}|p{0.9\textwidth}}
\toprule
Task Type & Title Generation \\ \hline
Task ID & task418\_persent\_title\_generation \\ \hline
Definition &
  Given a document, generate a short title of the document. The title should convey the main idea/event/topic about which the document is being written. Note that URLs in the text have been replaced with [Link].
   \\ \hline  
Positive Example & 
  \textbf{Input}: In a letter posted on the White House web site Tuesday  John Brennan  assistant to President Barack Obama for homeland security and counterterrorism  said Schmidt will have regular access to the president and play a vital role in the country's security. Schmidt's selection comes more than 10 months after Obama declared cyber security a priority and ordered a broad administration review. A senior White House official said Obama was personally involved in the selection process and chose Schmidt because of his unique background and skills. Schmidt will have regular and direct access to the president for cyber security issues  the official said. The official spoke on the condition of anonymity to discuss the selection process. At the same time  cyber experts and potential job candidates have complained that the position lacks the budgetary and policy-making authority needed to be successful. Schmidt will report to the National Security Council and closely support the National Economic Council on cyber issues. Schmidt's selection suggests that economic and business interests in the White House held more sway in the selection process. Schmidt  president and CEO of the Information Security Forum  a nonprofit international consortium that conducts research in information security  has served as chief security officer for Microsoft and as cyber security chief for online auction giant eBay. He was reportedly preferred by Lawrence Summers  director of the economic council. A good format for the titel can be the simple subject + object + verb. \newline 
  \textbf{Output}: White House picks new cyber coordinator \newline 
  \textbf{Explanation}: The title is relevant to the main topic of document, that is, the selection of Schmidt as the cybersecutiy chief.
   \\ \hline 
Negative Example & 
  \textbf{Input}: Lauren Cohan's Walking Dead fate may be written in the Whiskey Cavalier. While the show hasn't been picked up yet   Cohan  hasn't yet signed on for season 9 of Walking Dead and rumors have circulated for months about her possible exit from the zombie show. Just two days ago show runner Scott Gimple told TV Line that  he is incredibly hopeful about working out a new deal with  Cohan  who has played Maggie since the second season. ``These things do happen in TV'' he  said. But we are talking. \newline
  \textbf{Output}: Scott Gimple signed Lauren Cohan for season 9
  \newline 
  \textbf{Explanation}: Though the topic is relevant  and have correct names from the document, this is incorrect fact, as it is mentioned that Cohan hasn't yet signed.
   \\ \hline
Instance & 
  \textbf{Input}: Days after at least 58 people were killed in a Las Vegas mass shooting , Hillary Clinton called for better gun control . Clinton also had some words for President Trump , particularly of his handling of Hurricane Maria and the devastation in Puerto Rico . Clinton , on her book tour for ``What Happened ,'' called her memoir ``a story of resilience .'' Fallon also had female staff writers write thank you notes to Clinton . ``Thank you , Miley , tonight 's show writers and all of the women and young girls out there who are smart , strong and deserving of every opportunity ,'' Clinton said .
 As for election night , Clinton said she was disappointed both that she lost and that President Trump won . \newline 
  \textbf{Valid Output}: [``Hillary Clinton calls for gun control after Las Vegas shooting'']
  \\
 \bottomrule
\end{tabular}%
}
\caption{An example task in the Title Generation category of our dataset, adopted from PerSenT~\cite{bastan2020author}.}
\label{example:title_generation}
\end{table*}

\begin{table*}[htbp]
\centering
\small
\resizebox{0.9\textwidth}{!}{%
\begin{tabular}{p{0.1\textwidth}|p{0.9\textwidth}}
\toprule
Task Type & Data to Text \\ \hline
Task ID & task957\_e2e\_data\_to\_text \\ \hline
Definition &
  In this task, we ask you convert a data table of restaurant descriptions into fluent natural-sounding English sentences. The input is a string of key-value pairs; the output should be a natural and grammatical English sentence containing all the information from the input.
   \\ \hline  
Positive Example & 
  \textbf{Input}: name[Aromi], eatType[restaurant], food[English], area[city centre] \newline 
  \textbf{Output}: Aromi is an English restaurant in the city centre. \newline 
  \textbf{Explanation}: The output sentence faithfully converts the data in the input into a natural-sounding sentence.
   \\ \hline 
Negative Example & 
  \textbf{Input}: name[Blue Spice], eatType[coffee shop], priceRange[more than \u00a330], customer rating[5 out of 5], area[riverside], familyFriendly[yes], near[Avalon] \newline
  \textbf{Output}: Blue Spice is a Colombian coffee shop located by the riverside, near Avalon in Boston. Its prices are over \u00a330. Its customer ratings are 5 out of 5.
  \newline 
  \textbf{Explanation}: While the output contains most of the information from the input, it hallucinates by adding ungrounded information such as ``Colombian'' and ``Boston''.
   \\ \hline
Instance & 
  \textbf{Input}: name[The Mill], eatType[restaurant], area[riverside], near[The Rice Boat] \newline 
  \textbf{Valid Output}: [``A restaurant called The Mill, can be found near the riverside next to The Rice Boat.'']
  \\
 \bottomrule
\end{tabular}%
}
\caption{An example task in the Data to Text category of our dataset, adopted from E2E~\cite{novikova2017e2e}.}.
\label{example:data_to_text}
\end{table*}

\begin{table*}[ht]
\centering
\small
\resizebox{0.9\textwidth}{!}{%
\begin{tabular}{p{0.1\textwidth}|p{0.9\textwidth}}
\toprule
Task Type &  Grammar Error Correction \\ \hline
Task ID & task1557\_jfleg\_grammar\_error\_correction \\ \hline
Definition &
  In this task, you will be shown an incorrect English sentence. You need to generate a corrected form of the input sentence.
   \\ \hline  
Positive Example & 
  \textbf{Input}: The car's wheel are loose. \newline 
  \textbf{Output}: The car's wheel is loose. \newline 
  \textbf{Explanation}: The instance of are is replaced by the word is. This makes the sentence grammatically correct.
   \\ \hline 
Negative Example & 
  \textbf{Input}: This way is the way to go. \newline
  \textbf{Output}: This way may be the way to go.
  \newline 
  \textbf{Explanation}: The example does not correct the misuse of the word way. Instead, it should shorten the sentence to: this is the way to go.
   \\ \hline
Instance & 
  \textbf{Input}: I think it 's harder for successful preson to risk somethnig , thay coluld lost much more then others . \newline 
  \textbf{Valid Output}: [``I think it 's harder for a successful person to risk something becuase they could lose much more than others .'']
  \\
 \bottomrule
\end{tabular}%
}
\caption{An example task in the Grammar Error Correction category of our dataset, adopted from JFLEG~\cite{napoles2017jfleg}.}.
\label{example:grammar_error_correction}
\end{table*}

\section{Performance Improvement per Evaluation Task}
\label{sec:per-task performance}

\begin{figure*}[t!]
    \centering
    \includegraphics[width=\textwidth]{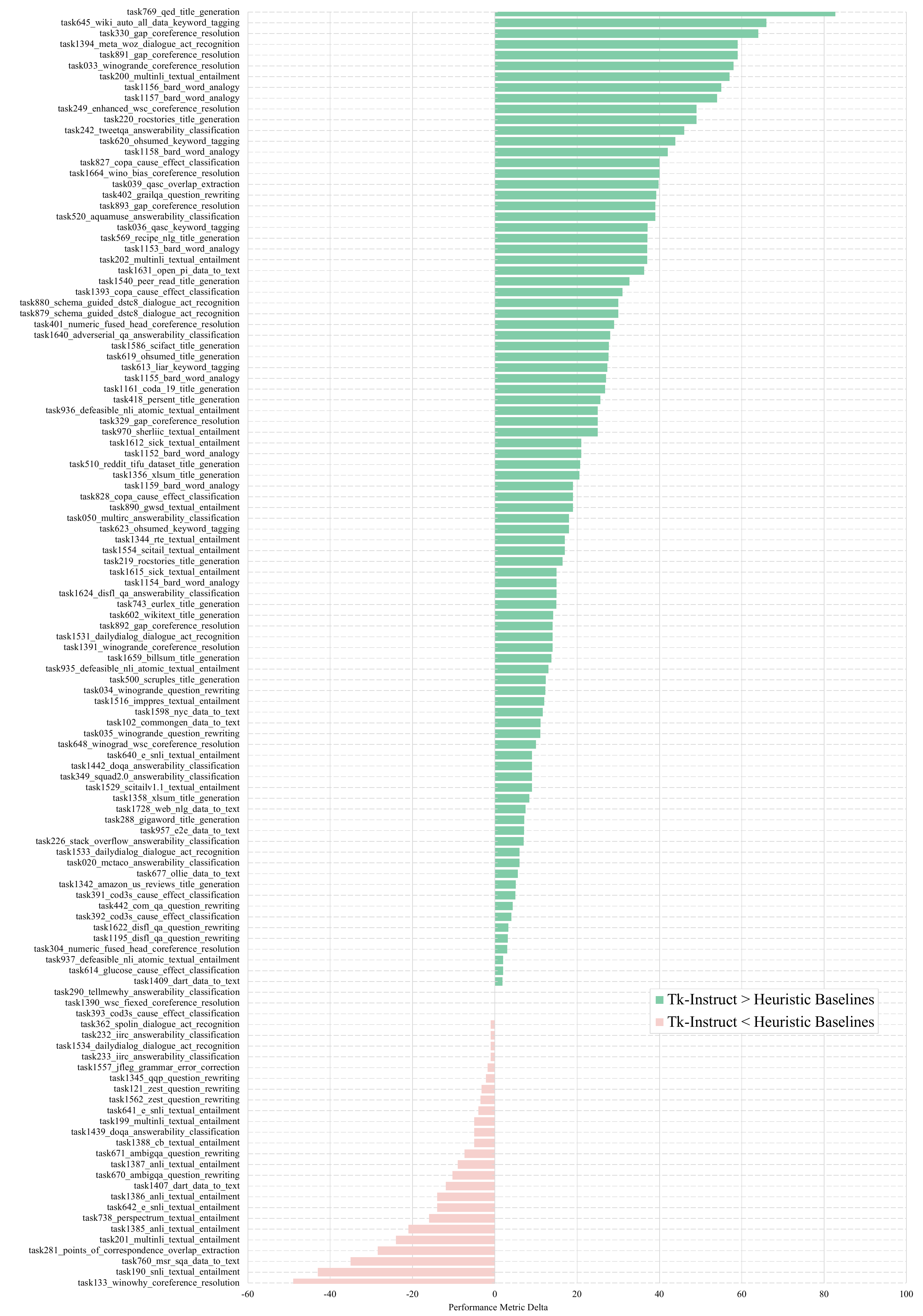}
    \caption{\modelName{}'s per-task performance improvement over the best of two heuristic baselines on the 119 evaluation tasks of the English track.}
    \label{fig:perf_improve_per_task}
\end{figure*}

\begin{figure*}[t!]
    \centering
    \includegraphics[width=\textwidth]{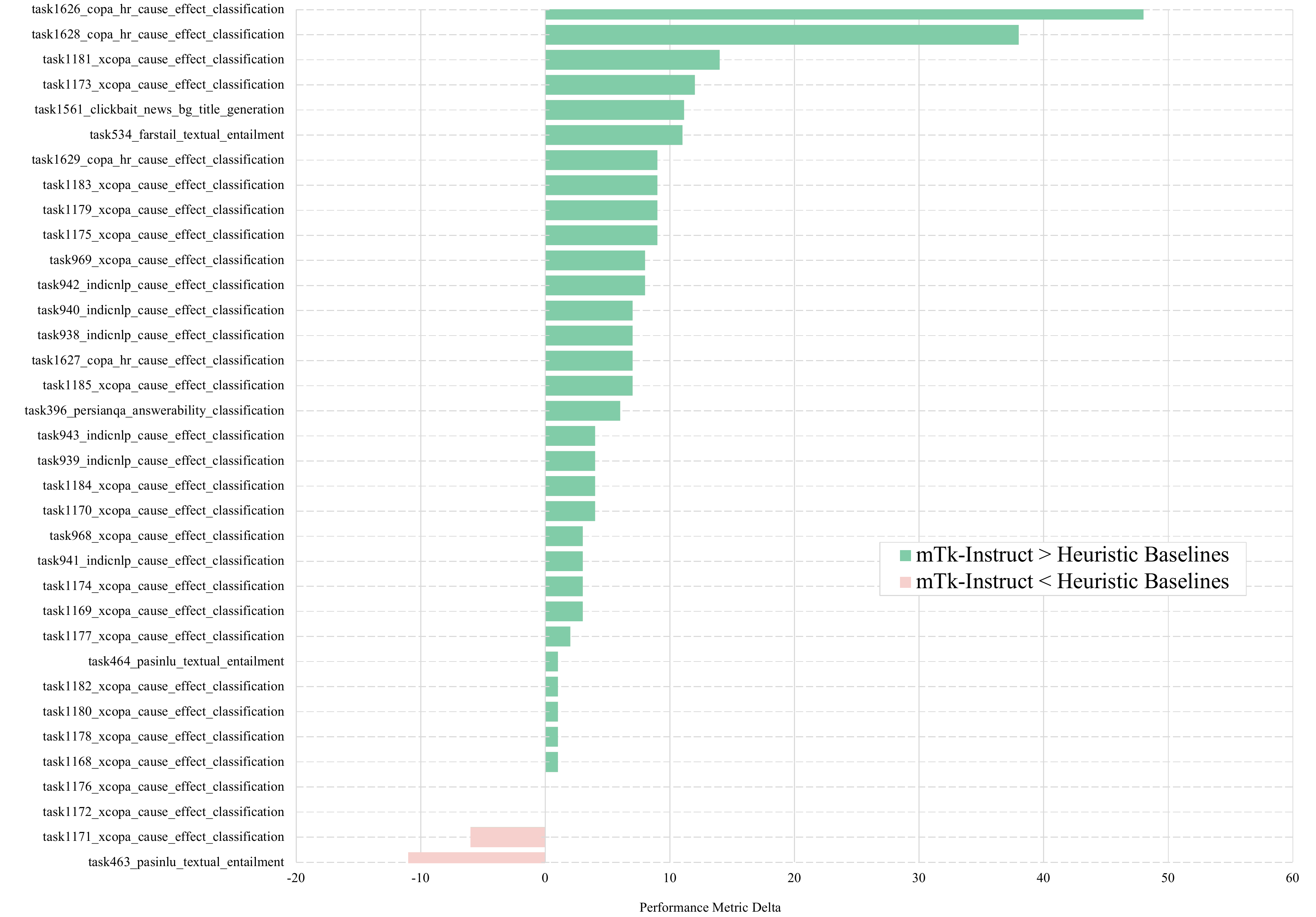}
    \caption{\modelName{}'s per-task performance improvement over the best of two heuristic baselines on the 35 evaluation tasks of the cross-lingual track.}
    \label{fig:perf_improve_per_task_non_en}
\end{figure*}

To provide more detailed analysis of \modelName{} on each individual task, Fig.~\ref{fig:perf_improve_per_task} presents the per-task improvement of our \modelName{} (3B) model over the best of two heuristic baselines on the English evaluation tasks, and Fig. \ref{fig:perf_improve_per_task_non_en} presents the per-task improvement of the m\modelName{} model on the cross-lingual evaluation tasks. For most of the evaluation tasks, we see a notable extent of generalization by \modelName{}.

\end{document}